\documentclass{article} 
\usepackage{iclr2023_conference,times}

\usepackage{amsmath,amsfonts,bm}

\def\eqref#1{equation~\ref{#1}}

\def\1{\bm{1}}

\DeclareMathAlphabet{\mathsfit}{\encodingdefault}{\sfdefault}{m}{sl}
\SetMathAlphabet{\mathsfit}{bold}{\encodingdefault}{\sfdefault}{bx}{n}

\usepackage{graphicx}
\usepackage{hyperref}
\usepackage{url}
\usepackage{tikz}
\usepackage{comment}
\usepackage{amsmath,amssymb} 
\usepackage{color}
\usepackage{multirow}

\usepackage{soul}  

\usepackage[normalem]{ulem}
\useunder{\uline}{\ul}{}

\usepackage{pbox}

\usepackage{amssymb}
\usepackage{wrapfig}
\usepackage{algorithm}
\usepackage[noend]{algcompatible}
\usepackage[accsupp]{axessibility}

\usepackage{wrapfig}
\usepackage{bbm}
\usepackage{enumitem} 

\usepackage{pdfpages}

\let\OLDthebibliography\thebibliography
\renewcommand\thebibliography[1]{
  \OLDthebibliography{#1}
  \setlength{\parskip}{2pt}
  \setlength{\itemsep}{2pt plus 1.3ex}
}

\newcommand{\nickname}{DM-NeRF}

\title{\nickname{}: 3D Scene Geometry Decomposition \\ and Manipulation from 2D Images}

\author{Bing Wang\textsuperscript{1,2,3}$^\dag$\quad Lu Chen\textsuperscript{1,2}$^\dag$\quad Bo Yang\textsuperscript{1,2}\thanks{Corresponding Author\quad \textsuperscript{\dag} Equal Contribution} \\
\textsuperscript{1} Shenzhen Research Institute, The Hong Kong Polytechnic University \\
\textsuperscript{2} vLAR Group, The Hong Kong Polytechnic University\quad \textsuperscript{3}University of Oxford\\
\texttt{bingwang@polyu.edu.hk\quad 
bo.yang@polyu.edu.hk} \\
}

\iclrfinalcopy 
\begin{document}

\maketitle

\begin{abstract}
In this paper, we study the problem of 3D scene geometry decomposition and manipulation from 2D views. By leveraging the recent implicit neural representation techniques, particularly the appealing neural radiance fields, we introduce an object field component to learn unique codes for all individual objects in 3D space only from 2D supervision. The key to this component is multiple carefully designed loss functions to enable every 3D point, especially in non-occupied space, to be effectively optimized without 3D labels. In addition, we introduce an inverse query algorithm to freely manipulate any specified 3D object shape in the learned scene representation. Notably, our manipulation algorithm can explicitly tackle key issues such as object collisions and visual occlusions. Our method, called \nickname{}, is among the first to simultaneously reconstruct, decompose, manipulate and render complex 3D scenes in a single pipeline. Extensive experiments on three datasets clearly show that our method can accurately decompose all 3D objects from 2D views, allowing any interested object to be freely manipulated in 3D space such as translation, rotation, size adjustment, and deformation.  
\end{abstract}

 \section{Introduction}
 In many cutting-edge applications such as mixed reality on mobile devices, users may desire to virtually manipulate objects in 3D scenes, such as moving a chair or making a flying broomstick in a 3D room. This would allow users to easily edit real scenes at fingertips and view objects from new perspectives. However, this is particularly challenging as it involves 3D scene reconstruction, decomposition, manipulation, and photorealistic rendering in a single framework \citep{Savva2019}. 

\vspace{-0.05cm}
A traditional pipeline firstly reconstructs explicit 3D structures such as point clouds or polygonal meshes using SfM/SLAM techniques \citep{Ozyesil2017,Cadena2016}, and then identifies 3D objects followed by manual editing. However, these explicit 3D representations inherently discretize continuous surfaces, and changing the shapes often requires additional repair procedures such as remeshing \citep{Alliez2002}. Such discretized and manipulated 3D structures can hardly retain geometry and appearance details, resulting in the generated novel views to be unappealing. Given this, it is worthwhile to design a new pipeline which can recover continuous 3D scene geometry only from 2D views and enable object decomposition and manipulation. 

\vspace{-0.05cm}
Recently, implicit representations, especially NeRF \citep{Mildenhall2020}, emerge as a promising tool to represent continuous 3D geometries from images. A series of succeeding methods \citep{Boss2021,Chen2021a,Zhang2021d} are rapidly developed to decouple lighting factors from structures, allowing free edits of illumination and materials. However, they fail to decompose 3D scene geometries into individual objects. Therefore, it is hard to manipulate individual object shapes in complex scenes. Recent works \citep{Stelzner2021,Zhang2021e,Kania2022,Yuan2022a,Tschernezki2022,Kobayashi2022,Kim2022,Benaim2022,Ren2022} have started to learn disentangled shape representations for potential geometry manipulation. However, they either focus on synthetic scenes or simple model segmentation, and can hardly extend to real-world 3D scenes with dozens of objects.

\begin{figure}[t]
\centering
   \includegraphics[width=.85\linewidth]{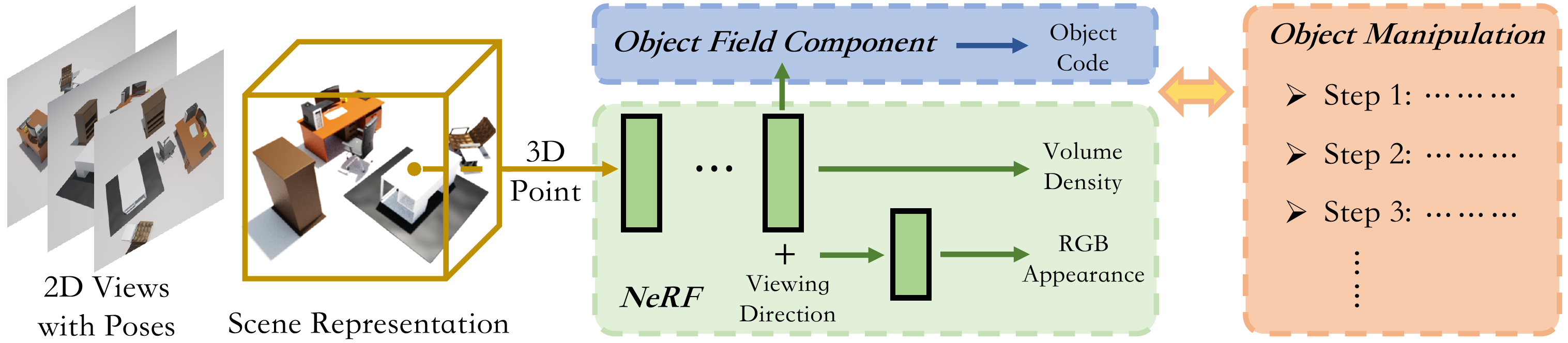}
   \vspace{-0.3cm}
\caption{The general workflow of \nickname{}. NeRF (green block) is used as the backbone. We propose the object field and manipulation components as illustrated by blue and orange blocks.}
\label{fig:opening}
\vspace{-.6cm}
\end{figure}

In this paper, we aim to simultaneously recover continuous 3D scene geometry, segment all individual objects in 3D space, and support flexible object shape manipulation such as translation, rotation, size adjustment and deformation. In addition, the edited 3D scenes can be also rendered from novel views. However, this task is extremely challenging as it requires: 1) an object decomposition approach amenable to continuous and implicit 3D fields, without relying on any 3D labels for supervision due to the infeasibility of collecting labels in continuous 3D space; 2) an object manipulation method agreeable to the learned implicit and decomposed fields, with an ability to clearly address visual occlusions inevitably caused by manipulation. 

To tackle these challenges, we design a simple pipeline, \textbf{\nickname{}}, which is built on the successful NeRF, but able to \textbf{d}ecompose the entire 3D space into object fields and freely \textbf{m}anipulate their geometries for realistic novel view rendering. As shown in Figure \ref{fig:opening}, it consists of 3 major components: 1) the existing radiance field to learn volume density and appearance for every 3D point in space; 2) the object field which learns a unique object code for every 3D point; 3) the object manipulator that directly edits the shape of any specified object and automatically tackles visual occlusions. 

The \textbf{object field} is the core of \nickname{}. This component aims to predict a one-hot vector, \textit{i.e.,} object code, for every 3D point in the entire scene space. However, learning such code involves critical issues: 1) there are no ground truth 3D object codes available for full supervision; 2) the number of total objects is variable and there is no fixed order for objects; 3) the non-occupied (empty) 3D space must be taken into account, but there are no labels for supervision as well. In Section \ref{sec:meth_objfield}, we show that our object field together with multiple carefully designed loss functions can address them properly, under the supervision of color images with 2D object masks only. 

Once the object field is well learned, our \textbf{object manipulator} aims to directly edit the geometry and render novel views when specifying the target objects, viewing angels, and manipulation settings. A na\"ive method is to obtain explicit 3D structures followed by manual editing and rendering, so that any shape occlusion and collision can be explicitly addressed. However, it is extremely inefficient to evaluate dense 3D points from implicit fields. To this end, as detailed in Section \ref{sec:meth_objmani}, we introduce a lightweight inverse query algorithm to automatically edit the scene geometry.

Overall, our pipeline can simultaneously recover 3D scene geometry, decompose and manipulate object instances only from 2D images. Extensive experiments on multiple datasets demonstrate that our method can precisely segment all 3D objects and effectively edit 3D scene geometry, without sacrificing high fidelity of novel view rendering. Our key contributions are:
\vspace{-0.2cm}
\begin{enumerate}[itemsep = -.5mm, leftmargin= 10 pt]
\item[$\bullet$] We propose an object field to directly learn a unique code for each object in 3D space only from 2D images, showing remarkable robustness and accuracy over the commonly-used individual image based segmentation methods. 

\item[$\bullet$] We propose an inverse query algorithm to effectively edit specified object shapes, while generating realistic scene images from novel views.

\item[$\bullet$] We demonstrate superior performance for 3D decomposition and manipulation, and also contribute the first synthetic dataset for quantitative evaluation of 3D scene editing. Our code and dataset are available at \url{https://github.com/vLAR-group/DM-NeRF}
\end{enumerate}
\vspace{-0.2cm}

We note that recent works ObjectNeRF \citep{Yang2021}, NSG \citep{Ost2021} and ObjectSDF \citep{Wu2022a} address the similar task as ours. However, ObjectNeRF only decomposes a foreground object, NSG focuses on decomposing dynamic objects, and ObjectSDF only uses semantic label as regularization. None of them directly learns to segment multiple 3D objects as ours. A few works \citep{Norman2022,Kundu2022,Fu2022} tackle panoptic segmentation in radiance fields. However, they fundamentally segment objects in 2D images followed by learning a separate radiance field for each object. By comparison, our method learns to directly segment all objects in the 3D scene radiance space, and it demonstrates superior accuracy and robustness than 2D object segmentation methods such as MaskRCNN \citep{He2017a} and Swin Transformer \citep{Liu2021SwinWindows}, especially when 2D object labels are noisy during training, as detailed in Section  \ref{sec:exp}.


\vspace{-0.3cm}

 \section{Related Work}
 \vspace{-0.25cm}
\textbf{Explicit 3D Representations:} 
To represent 3D geometry of objects and scenes, voxel grids \citep{Chan2016}, octree \citep{Tatarchenko2017}, meshes \citep{Kato2017,Groueix2018}, point clouds \citep{Fan2017} and shape primitives \citep{Zou2017} are widely used. Although impressive progress has been achieved in shape reconstruction \citep{Yang2018,Xie2019}, completion \citep{Song2017}, generation \citep{Lin2017a}, and scene understanding \citep{Tulsiani2017c,Gkioxari2019}, the quality of these representations are inherently limited by the spatial resolution and memory footprint. Therefore, they are hard to represent complex 3D scenes. 

\vspace{-0.1cm}
\textbf{Implicit 3D Representations:} 
To overcome the discretization issue of explicit representations, coordinate based MLPs have been recently proposed to learn implicit functions to represent continuous 3D shapes. These implicit representations can be generally categorized as: 1) signed distance fields \citep{Park2019}, 2) occupancy fields \citep{Mescheder2019}, 3) unsigned distance fields \citep{Chibane2020a,Wang2022}, 4) radiance fields \citep{Mildenhall2020}, and 5) hybrid fields \citep{Wang2021d}. Among them, both occupancy fields and signed distance fields can only recover closed 3D shapes, and are hard to represent open geometries. These representations have been extensively studied for novel view synthesis \citep{Niemeyer2019,Trevithick2021} and 3D scene understanding \citep{Zhang2021b,Zhi2021,Zhi2021a}. Thanks to the powerful representation, impressive results have been achieved, especially from the neural radiance fields and its succeeding methods. In this paper, we also leverage the success of implicit representations, particularly NeRF, to recover the geometry and appearance of 3D scenes from 2D images. 

\textbf{3D Object Segmentation:} 
To identify 3D objects from complex scenes, existing methods generally include 1) image based 3D object detection \citep{Mousavian2017}, 2) 3D voxel based detection methods \citep{Zhou2018a} and 3) 3D point cloud based object segmentation methods \citep{Yang2019d}. Given large-scale datasets with full 3D object annotations, these approaches have achieved excellent object segmentation accuracy. However, they are particularly designed to process explicit and discrete 3D geometries. Therefore, they are unable to segment continuous and fine-grained shapes, and fail to support geometry manipulation and realistic rendering. With the fast development of implicit representation, it is desirable to learn object segmentation for implicit surfaces. To the best of our knowledge, this paper is among the first to segment all 3D objects of implicit representations for complex scenes, only with color images and 2D object labels for supervision. 

\textbf{3D Scene Editing:} 
Existing methods of editing 3D scenes from images can be categorized as 1) appearance editing and 2) shape editing. A majority of works \citep{Sengupta2019,Boss2021,Zhang2021d,Chen2021a} focus on lighting decomposition for appearance editing. Although achieving appealing results, they cannot separately manipulate individual objects. A number of recent works \citep{Munkberg2021,Liu2021b,Jang2021,Stelzner2021,GuandaoYang2021} start to learn disentangled shape representations for potential geometry manipulation. However, they can only deal with single objects or simple scenes, without being able to learn unique object codes for precise shape manipulation and novel view rendering. In addition, there are also a plethora of works \citep{Tewari2020,Niemeyer2021,Dhamo2021,Alaluf2022} on generation based scene editing. Although they can manipulate the synthesized objects and scenes, they cannot discover and edit objects from real-world images. 

\vspace{-0.15cm}

 \section{\nickname{}}
 
\vspace{-0.05cm}
Given a set of $L$ images for a static scene with known camera poses and intrinsics { $\{(\mathcal{I}_1, \boldsymbol{\xi}_1, \boldsymbol{K}_1) \cdots (\mathcal{I}_L, \boldsymbol{\xi}_L, \boldsymbol{K}_L) \}$}, NeRF uses simple MLPs to learn the continuous 3D scene geometry and appearance. In particular, it takes 5D vectors of query point coordinates $\boldsymbol{p} = (x, y, z)$ and viewing directions $\boldsymbol{v} = (\theta, \phi)$ as input, and predicts the volume density $\sigma$ and color $\boldsymbol{c} = (r, g, b)$ for point $\boldsymbol{p}$. In our pipeline, we leverage this vanilla NeRF as the backbone to learn continuous scene representations, although other NeRF variants can also be used. Our method aims to decompose all individual 3D objects, and freely manipulate any object in the 3D scene space. To achieve this, we design an object field component to parallelly learn an object code for every query point $\boldsymbol{p}$, together with an object manipulator to edit the learned radiance fields and object fields. 

\begin{figure}[t]
\centering
   \includegraphics[width=.85\linewidth]{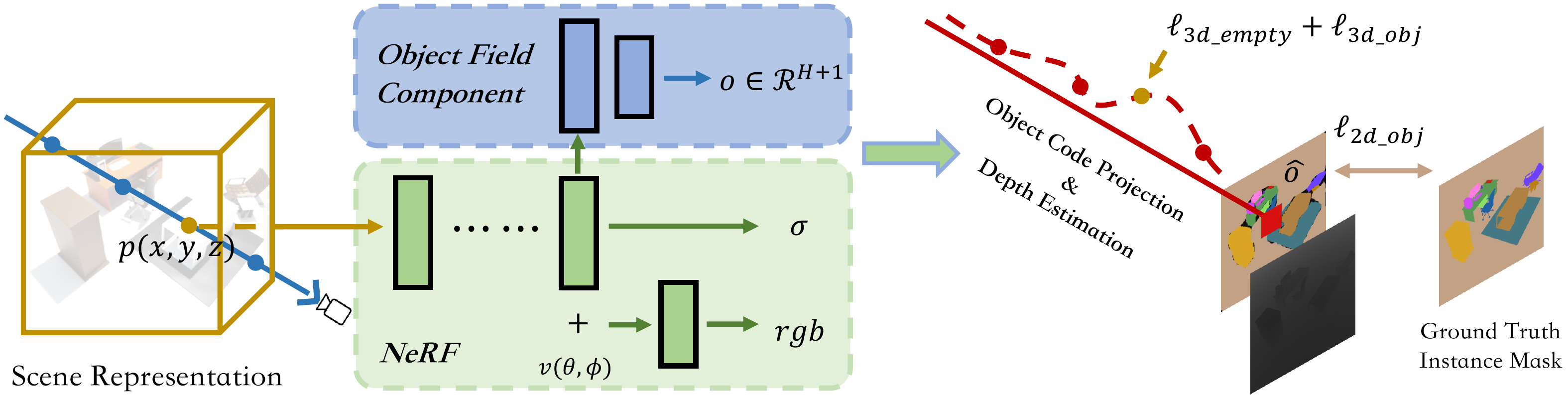}
   \vspace{-0.3cm}
\caption{The architecture of our pipeline. Given a 3D point $\boldsymbol{p}$, we learn an object code through three loss functions: $\ell_{2d\_obj}$/$\ell_{3d\_empty}$/$\ell_{3d\_obj}$ in Eqs \ref{eq:loss_2d_obj}\&\ref{eq:loss_3demp}\&\ref{eq:loss_3dobj}, using 2D and 3D supervision signals.}
\label{fig:meth_objfield}
\vspace{-0.5cm}
\end{figure}

\vspace{-0.15cm}
\subsection{Object Fields}\label{sec:meth_objfield}
\vspace{-0.15cm}
\textbf{Object Field Representation:} As shown in Figure \ref{fig:meth_objfield}, given the input point $\boldsymbol{p}$, we model the object field as a function of its coordinates, because the object signature of a 3D point is irrelevant to the viewing angles. The object field is represented by a one-hot vector $\boldsymbol{o}$. Basically, this one-hot object code aims to accurately describe the object ownership of any point in 3D space. 

However, there are two issues here: 1) the total number of objects in 3D scenes is variable and it can be 1 or many; 2) the entire 3D space has a large non-occupied volume in addition to solid objects. To tackle these issues, we define the object code $\boldsymbol{o}$ as $H+1$ dimensional, where $H$ is a predefined number of solid objects that the network is expected to predict in maximum. We can safely choose a relative large value for $H$ in practice. The last dimension of $\boldsymbol{o}$ is particularly reserved to represent the non-occupied space. Notably, this dedicated design is crucial for tackling occlusion and collision during object manipulation discussed in Section \ref{sec:meth_objmani}. Formally, the object field is defined as:
\vspace{-0.13cm}
\begin{equation}
\vspace{-0.27cm}
    \boldsymbol{o} = f(\boldsymbol{p}), \quad \text{where } \boldsymbol{o} \in \mathcal{R}^{H+1} 
\end{equation}

The function $f$ is parameterized by a series of MLPs. If the last dimension of code $\boldsymbol{o}$ is 1, it represents the input point $\boldsymbol{p}$ is non-occupied or the point is empty.

\textbf{Object Code Projection:} Considering that it is infeasible to collect object code labels in continuous 3D space for full supervision while it is fairly easy and low-cost to collect object labels on 2D images, we aim to project the object codes along the query light ray back to a 2D pixel. Since the volume density $\sigma$ learned by the backbone NeRF represents the geometry distribution, we simply approximate the projected object code of a pixel $\boldsymbol{\hat{o}}$ using the sampling strategy and volume rendering formulation of NeRF. Formally, it is defined as:
\vspace{-0.19cm}
\begin{equation}\label{eq:objcode_2d_proj}
\vspace{-0.16cm}
    \boldsymbol{\hat{o}} = \sum_{k=1}^K T_k \alpha_k \boldsymbol{o}_k, \quad \text{where} \quad T_k = exp(-\sum_{i=1}^{k-1}\sigma_i\delta_i), \quad \alpha_k = 1- exp(-\sigma_k\delta_k)
\end{equation}
with $K$ representing the total sample points along the light ray shooting from a pixel, $\sigma_i$ representing the learned density of the $i^{th}$ sample point, {$\delta_k$} representing the distance between the $(k+1)^{th}$ and $k^{th}$ sample points. From this projection formulation, we can easily obtain 2D masks of 3D object codes given the pose and camera parameters of any query viewing angles.


\setlength{\columnsep}{10pt}
\begin{wrapfigure}[9]{R}{0.34\textwidth}
\centering
\raisebox{5pt}[\dimexpr\height-.8\baselineskip\relax]{
\includegraphics[width=0.90\linewidth]{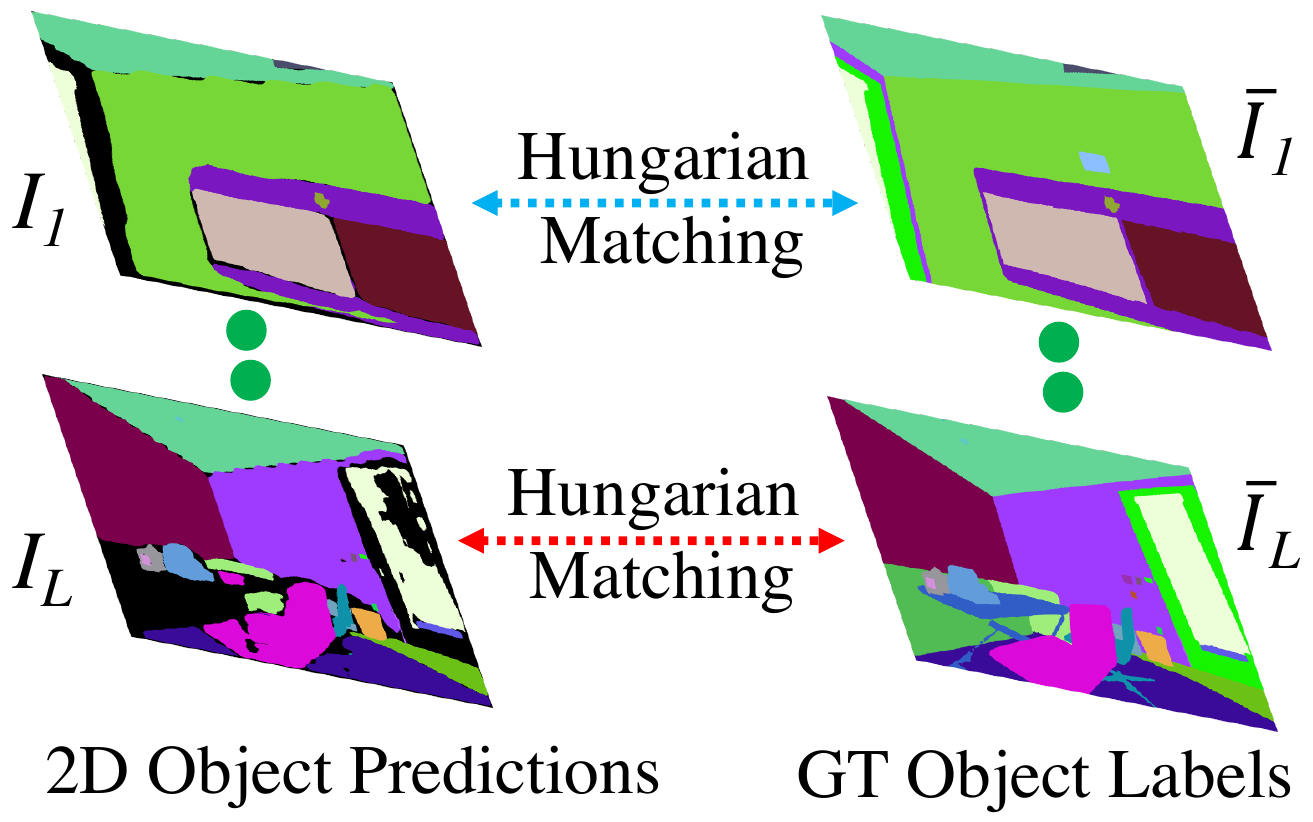}}
\vspace{-.55cm}
\caption{\small{Illustration of 2D object matching and supervision for $\ell_{2d\_obj}$.}}
\label{fig:dm-nerf_2dsup}
\end{wrapfigure}

\textbf{Object Code Supervision:} As shown in the right part of Figure \ref{fig:meth_objfield}, having projected 2D object predictions at hand, we choose 2D images with object annotations for supervision. However, there are two issues: 1) The number and order of ground truth objects can be very different across different views due to visual occlusions. For example, as to the same 3D object in space, its object annotation in image \#1 can be quite different from its annotation in image \#2. Therefore, it is non-trivial to consistently utilize 2D annotations for supervision. 2) The 2D annotations only provide labels for 3D solid objects, as non-occupied 3D space is never recorded in 2D images. Therefore, it is impossible to directly supervise non-occupied space, \textit{i.e.}, the last dimension of $\boldsymbol{\hat{o}}$, from 2D annotations. These issues make adaptions of existing 2D methods ineffective, \textit{e.g.}, Mask-RCNN \citep{He2017a} or Swin Transformer {\citep{Liu2021SwinWindows}}, fundamentally because they do not consider the consistency between 3D and 2D, but just segment objects on individual images.  

\textbf{\textit{To tackle the first issue}}, we use the Optimal Association and Supervision strategy proposed by 3D-BoNet \citep{Yang2019d}. As illustrated in Figure \ref{fig:dm-nerf_2dsup}, assuming we generate $L$ images of 2D object predictions $\{I_1 \dots I_l \dots I_L \}, I_l \in \mathcal{R}^{U \times V \times (H + 1)}$ and have the corresponding $L$ images of 2D ground truth object labels $\{\bar{I}_1 \dots \bar{I}_l \dots \bar{I}_L \}, \bar{I}_l \in \mathcal{R}^{U \times V \times T}$, in which $H$ is the predefined number of objects and $T$ represents the number of ground truth objects. Note that the ground truth number of objects in each image is usually different, but here we use the same $T$ to avoid an abuse of notation.

For each pair, we firstly take the first $H$ dimensions (solid object predictions) of $I$ and reshape them to be $M \in \mathcal{R}^{N \times H}$, where $N=U \times V$, while the last dimension is never used at this stage. Likewise, $\bar{I}$ is reshaped to be $\bar{M} \in \mathcal{R}^{N \times T}$. Then, $M$ and $\bar{M}$ are fed into Hungarian algorithm \citep{Kuhn1955} to associate every ground truth 2D object mask with a unique predicted 2D object mask, according to Soft Intersection-over-Union (sIoU) and Cross-Entropy Score (CES) \citep{Yang2019d}. Formally, the Soft Intersection-over-Union (sIoU) cost between the $h^{th}$ predicted object mask and the $t^{th}$ ground truth object mask in the $\l^{th}$ pair is defined as follows: 
\vspace{-0.1cm}
\begin{equation}
\setlength{\belowdisplayskip}{-2pt}
\boldsymbol{C}^{sIoU}_{h,t} = \frac{ - \sum_{n=1}^N(M^n_h * \bar{M}^n_t)}{ \sum_{n=1}^N M^n_h+\sum_{n=1}^{N}\bar{M}^n_t- \sum_{n=1}^{N}(M^n_h *\bar{M}^n_t)}
\vspace{-0.1cm}
\end{equation}

where $M^n_h$ and $\bar{M}^n_t$ are the $n^{th}$ values of $M_h$ and $\bar{M}_t$. The Cross-Entropy Score (CES) between $M_h$ and $\bar{M}_t$ is formally defined as:
\vspace{-0.2cm}
\begin{equation}
\setlength{\belowdisplayskip}{-2pt}
\boldsymbol{C}^{CES}_{h,t} =-\frac{1}{N} \sum_{n=1}^{N} \left[\bar{M}^n_t\log M^n_h + (1 - \bar{M}^n_t)\log(1-M^n_h)\right]
\vspace{-0.1cm}
\end{equation}

After the optimal association based on $(\boldsymbol{C}^{sIoU}_{h,t} + \boldsymbol{C}^{CES}_{h,t})$, we reorder the predicted object masks to align with the $T$ ground truth masks, and then we directly minimize the cost of all ground truth objects in every pair. The final loss $\ell_{2d\_obj}$ is defined by averaging across all $L$ image pairs.
\vspace{-.2cm}
\begin{equation}\label{eq:loss_2d_obj}
    \ell_{2d\_obj} = \frac{1}{L}\sum_{l=1}^L (sIoU_l + CES_l) \text{, where \phantom{x}} 
    sIoU_l = \frac{1}{T}\sum_{t=1}^T(\boldsymbol{C}^{sIoU}_{t,t})\text{,\phantom{x}} CES_l = \frac{1}{T}\sum_{t=1}^T(\boldsymbol{C}^{CES}_{t,t})
    \vspace{-.3cm}
\end{equation}



\textbf{\textit{To tackle the second issue}}, we turn to supervise the non-occupied object code in 3D space with the aid of estimated surface distances. In particular, given a specific query light ray on which we sample $K$ 3D points to compute the projected 2D object code $\boldsymbol{\hat{o}}$, we simultaneously compute an approximate distance $d$ between camera center and the surface hit point along that query light:
\vspace{-.2cm}
\begin{equation}
\vspace{-.3cm}
    d = \sum_{k=1}^K T_k \alpha_k \delta_k, \quad \text{where} \quad T_k = exp(-\sum_{i=1}^{k-1}\sigma_i\delta_i), \quad \alpha_k = 1- exp(-\sigma_k\delta_k)
\end{equation}
where $K$ represents the total sample points along light ray, $\sigma_i$ is the learned density, and {$\delta_k$} is the distance between $(k+1)^{th}$ and $k^{th}$ sample points, as same as Equation \ref{eq:objcode_2d_proj}.

\setlength{\columnsep}{10pt}
\begin{wrapfigure}[10]{R}{0.347\textwidth}
\centering
\raisebox{5pt}[\dimexpr\height-.9\baselineskip\relax]{
\includegraphics[width=0.85\linewidth]{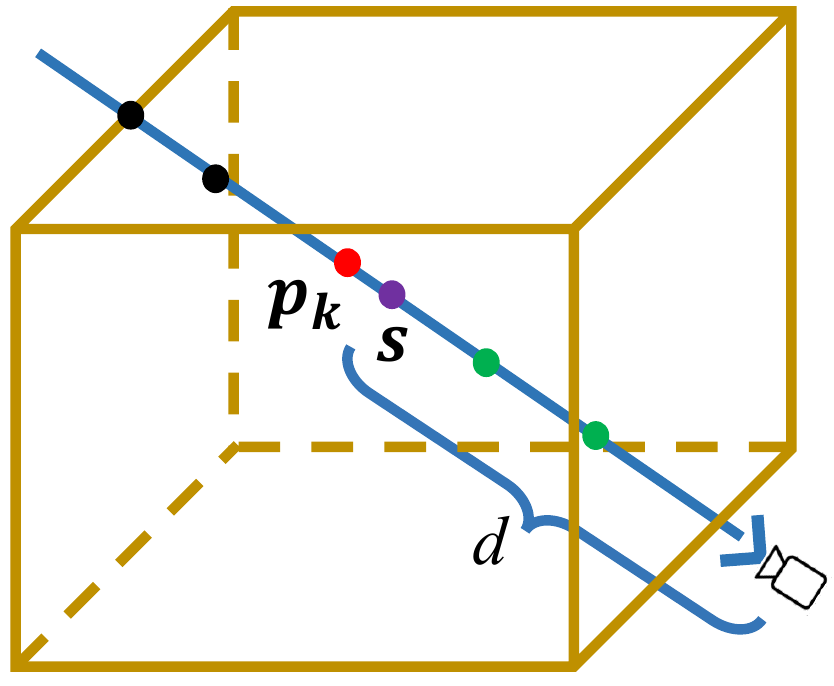}}
\vspace{-0.55cm}
\caption{\small {Empty points identification.}}
\label{fig:dm-nerf_3dsup}
\end{wrapfigure}
As shown in Figure \ref{fig:dm-nerf_3dsup}, once we have the surface distance $d$ at hand, we can easily know the relative position between every $k^{th}$ sample point and the surface point $s$ along the light ray. Naturally, we can then identify the subset of sample points surely belonging to empty space as indicated by green points, the subset of sample points near the surface as indicated by red points, and the remaining subset of sample points behind the surface as indicated by black points. Such geometric information provides critical signals to supervise empty space, \textit{i.e.}, the last dimension of object code $\boldsymbol{o}$. Note that, the sample points behind the surface may not surely belong to empty space, and therefore cannot be used as supervision signals. We use the following kernel functions to obtain a surfaceness score $s_k$ and an emptiness score $e_k$ for the $k^{th}$ sample point with the indicator function represented by $\mathbbm{1}()$.  
\begin{equation}\label{eq:scores_3demp}
    s_k = exp\big(-(d_k - d)^2\big), \quad e_k = (1 - s_k)* \mathbbm{1}(d - \Delta d - d_k > 0) 
\end{equation}
where $d_k$ represents the distance between camera center and the $k^{th}$ sample point, and $\Delta d$ is a hyperparameter to compensate the inaccuracy of estimated surface distance $d$. Note that, the indicator function is defined as: $\mathbbm{1}()=1$ if $d - \Delta d - d_k > 0$, and otherwise $\mathbbm{1}()=0$. It is used to mask out the sample points behind surface point during loss calculation. Given the emptiness and surfaceness scores for the total $K$ sample points along the ray, we  use the simple log loss to supervise the last dimension of object code denoted as $o^{H+1}_k$. 
\vspace{-.1cm}
\begin{equation}\label{eq:loss_3demp}
\vspace{-.2cm}
    \ell_{3d\_empty} = -\frac{1}{K}\sum_{k=1}^K \Big( e_k*log(o^{H+1}_k) + s_k*log(1-o^{H+1}_k) \Big)
\end{equation}

Since there should be no solid object at all in the empty space, we apply the following $\ell_{3d\_obj}$ loss on the first $H$ dimensions of the object code to push them to be zeros, being complementary to the existing $\ell_{2d\_obj}$ loss. The $h^{th}$ dimension of the $k^{th}$ sample point's object code is denoted by $o_k^h$. 
\vspace{-.2cm}
\begin{equation}\label{eq:loss_3dobj}
\vspace{-.3cm}
    \ell_{3d\_obj} = - \frac{1}{K} \sum_{k=1}^K  \Big( e_k*\sum_{h=1}^H log(1 - o_k^h)  \Big)
\end{equation}

To sum up, we firstly project object codes in 3D space back to 2D pixels along light rays using volume rendering equation, and then use optimal association strategy to compute the loss value with 2D object labels only. In addition, we introduce the key emptiness and surfaceness scores for points in 3D space with the aid of estimated surface distances. These unique scores are used to supervise the non-occupied 3D space. The whole object field is jointly supervised by:
\begin{equation}\label{eq:loss_obj}
    \ell = \ell_{2d\_obj} + \ell_{3d\_empty} + \ell_{3d\_obj}
\end{equation}

\vspace{-.5cm}
\subsection{Object Manipulator}\label{sec:meth_objmani}
\vspace{-.1cm}
If we want to manipulate a specific 3D object shape (\textit{e.g.,} translation, rotation and size adjustment), how does the whole scene look like in a new perspective after manipulation, assuming the object code and manipulation matrix are precomputed from users' interactions (\textit{e.g.,} click, drag or zoom)?

Intuitively, there could be two strategies: 1) firstly project 3D scene into 2D images, and then edit objects in 2D space. 2) firstly edit objects in 3D space, and then project into 2D images. Compared with the first strategy which would inevitably incur inconsistency across multiple images due to the independent edits on individual views, the latter is more favourable. Remarkably, our object field component can support the latter strategy. Regarding such a manipulation task on implicit fields, the core question is: how do we edit the codes $\sigma$/$\boldsymbol{c}$/$\boldsymbol{o}$ of every sample point along the query light ray, such that the generated novel view exactly shows  the new appearance? This is nontrivial as:
\vspace{-0.3cm}
\begin{enumerate}[itemsep = -.5mm, leftmargin= 10 pt]
    \item[$\bullet$] First, we need to address potential collisions between objects during manipulation. This is quite intuitive, thanks to our special design of emptiness score in the last dimension of object code $\boldsymbol{o}$. 
    \item[$\bullet$] Second, due to visual occlusions, object codes behind surface points may not be accurate as they are not sufficiently optimized. By comparison, the projected object code $\boldsymbol{\hat{o}}$ along a light ray tends to be more accurate primarily because we have ground truth 2D labels for strong supervision.
    \item[$\bullet$] At last, we need a systematic procedure to update the codes with the known manipulation information. To this end, we design an inverse query approach.
\end{enumerate}
\vspace{-0.3cm}

\setlength{\columnsep}{10pt}
\begin{wrapfigure}[9]{R}{0.346\textwidth}
\centering
\raisebox{5pt}[\dimexpr\height-1.3\baselineskip\relax]{
\includegraphics[width=0.85\linewidth]{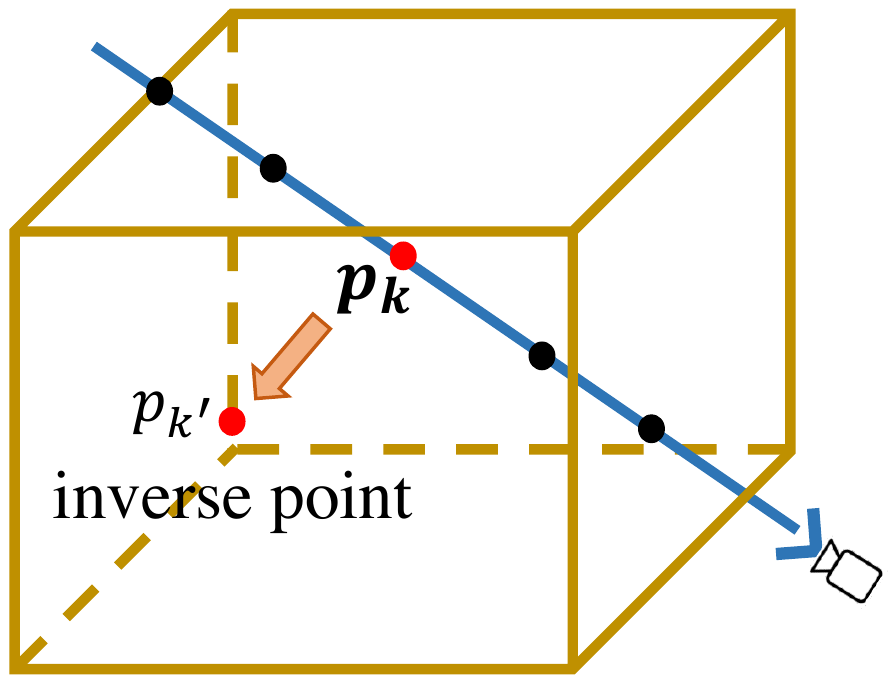}}
\vspace{-.55cm}
\caption{\small{Inverse points computation.}}
\label{fig:dm-nerf_edit}
\end{wrapfigure}

\textbf{Inverse Query:} We design a creative inverse query approach to address all above issues, realizing the favourable strategy: \textit{editing in 3D space followed by 2D projection}. In particular, as shown in Figure \ref{fig:dm-nerf_edit}, for any 3D sample point $\boldsymbol{p}_k$ along a specific query light ray, given the \textit{target} (\textit{i.e.}, to-be-edited) object code $\boldsymbol{o}_t$ and its manipulation settings: relative translation  $\Delta \boldsymbol{p}$$=$$(\Delta x, \Delta y, \Delta z)$, rotation matrix $\mathbf{R}^{3\times3}$, and scaling factor $t$$>$$0$, we firstly compute an inverse 3D point $\boldsymbol{p}_{k'}$, and then evaluate whether $\boldsymbol{p}_k$ and $\boldsymbol{p}_{k'}$ belong to the target object, and lastly decide to whether edit the codes or not. Formally, we introduce Inverse Query Algorithm \ref{alg:inv_query} to conduct a single light ray editing and rendering for object shape manipulation. Naturally, we can shoot a bunch of rays from any novel viewing angles to generate images of manipulated 3D scenes.

\begin{figure}[thb]
\vspace{-0.6cm}
\begin{algorithm}[H]
\caption{ {\small Our Inverse Query Algorithm to manipulate the learned implicit fields. (1) $\boldsymbol{o}_t$ is the target object code, one-hot with $H+1$ dimensions. $\{ \Delta \boldsymbol{p}, \mathbf{R}^{3\times3}, t > 0\}$ represent the manipulation information for the target object. (2) $\{ \boldsymbol{p}_1 \cdots \boldsymbol{p}_k \cdots \boldsymbol{p}_K \}$ represent the $K$ sample points along a specific query light ray $\boldsymbol{r}$. Note that, we convert all object codes into hard one-hot vectors for easy implementation.}
}
\label{alg:inv_query}
\begin{algorithmic} 
\footnotesize
\STATE{\textbf{Input:}} 
    \STATE{\phantom{xx}$\bullet$ The target object code $\boldsymbol{o}_t$, manipulation information $\{ \Delta \boldsymbol{p}, \mathbf{R}^{3\times3}, t\geq 0\}$; }
    \STATE{\phantom{xx}$\bullet$ The sample points $\{ \boldsymbol{p}_1 \cdots \boldsymbol{p}_k \cdots \boldsymbol{p}_K \}$ along a specific query light ray $\boldsymbol{r}$;}
\STATE{\textbf{Output:}} 
    \STATE{\phantom{xx}$\bullet$ The final pixel color $\boldsymbol{\bar{c}}$} rendered from the query ray $\boldsymbol{r}$ after manipulation;
    \STATE{\phantom{xx}$\bullet$ The final pixel object code $\boldsymbol{\bar{o}}$} rendered from the query ray $\boldsymbol{r}$ after manipulation;
    
\STATE{\textbf{Preliminary step:}} 
    \STATE{\phantom{xx}$\bullet$ Obtain the projected pixel object code $\boldsymbol{\hat{o}}$ of light ray $\boldsymbol{r}$ before manipulation;}

\STATE{\textit{NOTE: The loop below shows how to edit per sample point in 3D space.}}
\FOR {$\boldsymbol{p}_k$ in $\{\boldsymbol{p}_1 \cdots \boldsymbol{p}_k \cdots \boldsymbol{p}_K\}$ }{}
    \STATE{$\bullet$ Compute the inverse point $\boldsymbol{p}_{k'}$ for $\boldsymbol{p}_k$: \phantom{x}  
    $\boldsymbol{p}_{k'} = (1/t)\mathbf{R}^{-1}(\boldsymbol{p}_k - \Delta \boldsymbol{p})$;}
    \STATE{$\bullet$ Obtain the codes $\{\sigma_k$, $\boldsymbol{c}_k$, $\boldsymbol{o}_k\}$ for the point $\boldsymbol{p}_k$;}
    \STATE{$\bullet$ Obtain the codes $\{\sigma_{k'}$, $\boldsymbol{c}_{k'}$, $\boldsymbol{o}_{k'}\}$ for the inverse point $\boldsymbol{p}_{k'}$;}
    \STATE{$\bullet$ \textit{Tackle visual occlusions:}}
    \STATE{\phantom{xx} if $\boldsymbol{o}_k = \boldsymbol{o}_t$ and $\boldsymbol{o}_k \neq \boldsymbol{\hat{o}}$ do: \phantom{x} $\boldsymbol{o}_k \leftarrow \boldsymbol{\hat{o}} $ }
    \STATE{\phantom{xx} \textit{Note: the target object is behind the surface but will be manipulated; }}
    \STATE{$\bullet$ Obtain new implicit codes $\{\bar{\sigma}_k, \boldsymbol{\bar{c}}_k, \boldsymbol{\bar{o}}_k \}$ for $\boldsymbol{p}_k$ after manipulation:}
    \STATE{\phantom{xx} if $\boldsymbol{o}_k \neq \boldsymbol{o}_t$ and $o_k^{H+1} \neq 1$ and $\boldsymbol{o}_{k'} = \boldsymbol{o}_t$ do: \phantom{x} collision detected, EXIT.}
    \STATE{\phantom{xx} if $\boldsymbol{o}_k \neq \boldsymbol{o}_t$ and $\boldsymbol{o}_{k'} = \boldsymbol{o}_t$ do: 
          \phantom{x} $\{\bar{\sigma}_k, \boldsymbol{\bar{c}}_k, \boldsymbol{\bar{o}}_k\} \leftarrow \{\sigma_{k'}$, $\boldsymbol{c}_{k'}$, $\boldsymbol{o}_{k'}\} $ ;}
    \STATE{\phantom{xx} if $\boldsymbol{o}_k = \boldsymbol{o}_t$ and $\boldsymbol{o}_{k'} = \boldsymbol{o}_t$ do: 
          \phantom{x} $\{\bar{\sigma}_k, \boldsymbol{\bar{c}}_k, \boldsymbol{\bar{o}}_k \} \leftarrow \{\sigma_{k'}$, $\boldsymbol{c}_{k'}$, $\boldsymbol{o}_{k'}\} $ ;}
    \STATE{\phantom{xx} if $\boldsymbol{o}_k = \boldsymbol{o}_t$ and $\boldsymbol{o}_{k'} \neq \boldsymbol{o}_t$ do: 
          \phantom{x} $\{\bar{\sigma}_k, \boldsymbol{\bar{c}}_k, \boldsymbol{\bar{o}}_k \} \leftarrow \{ 0, \boldsymbol{0}, \boldsymbol{0} \}$ ;}
    \STATE{\phantom{xx} if $\boldsymbol{o}_k \neq \boldsymbol{o}_t$ and $\boldsymbol{o}_{k'} \neq \boldsymbol{o}_t$ do:
         \phantom{x} $\{\bar{\sigma}_k, \boldsymbol{\bar{c}}_k, \boldsymbol{\bar{o}}_k \} \leftarrow \{\sigma_k$, $\boldsymbol{c}_k$, $\boldsymbol{o}_k\} $ ;}
\ENDFOR
\\ After the above \textit{for loop}, every point $\boldsymbol{p}_k$ will get new implicit codes $\{\bar{\sigma}_k, \boldsymbol{\bar{c}}_k, \boldsymbol{\bar{o}}_k \}$. 

\STATE{\textit{NOTE: The step below shows how to project edited 3D points to a 2D image.}}\\
According to volume rendering equation, the final pixel color and object code are:
\STATE{\phantom{xx}$\bullet$}
    $\boldsymbol{\bar{c}} = \sum_{k=1}^K \bar{T}_k \bar{\alpha}_k \boldsymbol{\bar{c}}_k, \quad 
     \boldsymbol{\bar{o}} = \sum_{k=1}^K \bar{T}_k \bar{\alpha}_k \boldsymbol{\bar{o}}_k$ \\
    \phantom{xxx}$\textit{where} \quad \bar{T}_k = exp(-\sum_{i=1}^{k-1}\bar{\sigma}_i\delta_i), \quad \bar{\alpha}_k = 1- exp(-\bar{\sigma}_k\delta_k)$.

\end{algorithmic}
\end{algorithm}
\vspace{-.85cm}
\end{figure}

\vspace{-.2cm}
\subsection{Implementation}
\vspace{-.1cm}
To preserve the high-quality of image rendering, our loss in Equation \ref{eq:loss_obj} is only used to optimize the MLPs of object field branch. The backbone is only optimized by the original NeRF photo-metric loss \citep{Mildenhall2020}. The whole network is end-to-end trained from scratch. The single hyper-parameter for our object field $\Delta d$ is set as 0.05 meters in all experiments.
\vspace{-.3cm}

 \section{Experiments}\label{sec:exp}
 
\vspace{-.3cm}
\subsection{Datasets}
\vspace{-.15cm}
\textbf{DM-SR:} To the best of our knowledge, there is no 3D scene dataset suitable for quantitative evaluation of geometry manipulation. Therefore, we create a \textbf{s}ynthetic dataset with 8 different and complex indoor \textbf{r}ooms, called DM-SR. For each scene, we generate the following 5 groups of images:
\vspace{-.3cm}
\begin{enumerate}[itemsep = -.5mm, leftmargin= 10 pt]
    \item[$\bullet$] Group 1 (w/o Manipulation): Color images and 2D object masks at 400$\times$400 pixels are rendered from viewpoints on the upper hemisphere. We generate 300 views for training. 
    \item[$\bullet$] Group 2 (Translation Only): One object is selected to be translated along $x$ or $y$ axis with $\sim 0.3m$.
    \item[$\bullet$] Group 3 (Rotation Only): One object is selected to be rotated around $z$ axis with about 90 degrees.
    \item[$\bullet$] Group 4 (Scaling Only): One object is selected to be scaled down about 0.8$\times$ smaller.
    \item[$\bullet$] Group 5 (Joint Translation/Rotation/Scaling): One object is selected to be simultaneously translated about $\sim 0.3m$, rotated about 90 degrees, scaled down about 0.8$\times$ smaller.
\end{enumerate}
\vspace{-.3cm}
For each group, 100 views are generated for testing at the same viewpoints.

\textbf{Replica:} Replica \citep{Straub2019} is a reconstruction-based 3D dataset of high fidelity scenes. We request the authors of Semantic-NeRF \citep{Zhi2021} to generate (180 training / 180 testing) color images and 2D object masks with camera poses at 640$\times$480 pixels for each of 7 scenes. Each scene has 59$\sim$93 objects with very diverse sizes.

\textbf{ScanNet:} ScanNet \citep{Dai2017} is a large-scale challenging real-world dataset. We select 8 scenes ($\sim$10 objects in each one) for evaluation. Each scene has $\sim$3000 raw images with 2D object masks and camera poses, among which we evenly select 300 views for training and 100 for testing.

\vspace{-0.3cm}
\subsection{Baseline and Metrics} \label{sec:baselines}
\vspace{-0.3cm}
\textbf{Scene Decomposition:} The most relevant work to us is ObjectNeRF \citep{Yang2021}, but its design is vastly different: 1) It requires a point cloud as input for voxelization in training, but we do not. 2) It needs GT bounding boxes of target objects to manually prune point samples during editing, but we do not need any annotations in editing. 3) It only learns to binarily segment the foreground object and background, by pre-defining an Object Library in training and editing. However, our object code is completely learned from scratch. This means that ObjectNeRF is not comparable due to the fundamental differences. Note that, the recent Semantic-NeRF \citep{Zhi2021} is also not comparable because it only learns 3D semantic categories, not individual 3D objects. In fact, we notice that all recent published relevant works \citep{Yang2021,Zhi2021,Wu2022a,Kundu2022,Norman2022,Fu2022} do not directly tackle and evaluate multiple 3D object segmentation in literature. In this regard, we use the powerful Mask-RCNN \citep{He2017a} and Swin Transformer {\citep{Liu2021SwinWindows}} as baselines. 

Since we train our \nickname{} in scene-specific fashion, for fairness, we also fine-tune Mask-RCNN and Swin Transformer models (pretrained under Detectron2 Library) on every single scene. In particular, we carefully fine-tune both models using up to 480 epochs until convergence with learning rate $5e^{-4}$ and then pick up the best models on the testing split of each scene for comparison.

\textbf{Object Manipulation:} Since there is no method that can directly manipulate objects in continuous radiance fields, we adapt the recent Point-NeRF \citep{Xu2022Point-NeRF:Fields} for comparison, because it can recover both explicit 3D scene point clouds and novel views. Adaptation details are in Appendix.

\textbf{Metrics:} We use the standard PSNR/SSIM/LPIPS scores to evaluate color image synthesis \citep{Mildenhall2020}, and use AP of all 2D test images to evaluate 3D scene decomposition. 

\vspace{-.13cm}
\subsection{3D Scene Decomposition}\label{sec:exp_scene_decomp}
\vspace{-.23cm}
\textbf{Training with 100\% Accurate 2D Labels}: We evaluate the performance of scene decomposition on 3 datasets. For DM-SR dataset, we evaluate our method and baselines on images of Group 1 only, while the images of Groups 2/3/4/5 are used for object manipulation. For every single scene in these datasets, using 100\% accurate 2D object ground truth labels, we train a separate model for our method, and fine-tune separate models for Mask-RCNN (MR) and Swin Transformer (SwinT). 

\textbf{Analysis}: Table \ref{tbl:decomp_dm-sr} shows that our method, not surprisingly, achieves excellent results for novel view rendering thanks to the original NeRF backbone. Notably, our method obtains nearly perfect object segmentation results across multiple viewing points of complex 3D scenes in all three datasets, clearly outperforming baselines. Figure \ref{fig:qualitative_decomp} shows that our results have much sharper object boundaries thanks to the explicit 3D geometry applied in our object field.

\vspace{-.2cm}
\begin{table*}[h]\tabcolsep=0.13cm
\centering
\resizebox{1\linewidth}{!}{
\begin{tabular}{r|ccc|ccc|| r|ccc|ccc|| r|ccc|ccc}
\multicolumn{7}{c||}{DM-SR Dataset} & \multicolumn{7}{c||}{Replica Dataset} & \multicolumn{7}{c}{ScanNet Dataset} \\  \hline

 & \multicolumn{3}{c|}{\small{Novel View Synthesis}} & \multicolumn{3}{c||}{\small{Decomposition}}& & \multicolumn{3}{c|}{\small{Novel View Synthesis}} & \multicolumn{3}{c||}{\small{Decomposition}} & & \multicolumn{3}{c|}{\small{Novel View Synthesis}} & \multicolumn{3}{c}{\small{Decomposition}} \\ \hline

 & \small{PSNR$\uparrow$} & \small{SSIM$\uparrow$} & \small{LPIPS$\downarrow$} & \small{MR} & \small{SwinT} & \textbf{\small{Ours}} &  & \small{PSNR$\uparrow$} & \small{SSIM$\uparrow$} & \small{LPIPS$\downarrow$} & \small{MR} & \small{SwinT} & \textbf{\small{Ours}} & & \small{PSNR$\uparrow$} & \small{SSIM$\uparrow$} & \small{LPIPS$\downarrow$} & \small{MR} & \small{SwinT} & \textbf{\small{Ours}} \\
Bathroom  & 44.05   & 0.994   & 0.009   & 93.81 & 98.89 & 100.0 &             &         &        &         &        & &       & 0010\_00    & 26.82   & 0.809   & 0.381     & 83.90  & 87.59 & 94.82  \\
Bedroom   & 48.07   & 0.996   & 0.009   & 97.92 & 98.85 & 100.0 & Office\_0   & 40.66   & 0.972  & 0.070   & 74.05  & 80.17 & 82.71 & 0012\_00    & 29.28   & 0.753   & 0.389     & 86.90  & 89.92 & 98.86 \\
Dinning   & 42.34   & 0.984   & 0.028   & 98.85 & 97.81 & 99.66 & Office\_2   & 36.98   & 0.964  & 0.115   & 73.41  & 75.39 & 81.12 & 0024\_00    & 23.68   & 0.705   & 0.452     & 69.87  & 67.88 & 93.25 \\
Kitchen   & 46.06   & 0.994   & 0.014   & 92.04 & 98.81 & 100.0 & Office\_3   & 35.34   & 0.955  & 0.078   & 72.91  & 73.26 & 76.30 & 0033\_00    & 27.76   & 0.856   & 0.342     & 88.70  & 94.23 & 97.02 \\
Reception & 42.59   & 0.993   & 0.008   & 98.81 & 95.75 & 100.0 & Office\_4   & 32.95   & 0.921  & 0.172   & 74.76  & 72.51 & 70.33 & 0038\_00    & 29.36   & 0.716   & 0.415     & 96.01  & 97.94 & 99.17 \\
Rest      & 42.80   & 0.994   & 0.007   & 98.89 & 94.50 & 99.89 & Room\_0     & 34.97   & 0.940  & 0.127   & 78.67  & 76.90 & 79.83 & 0088\_00    & 29.37   & 0.825   & 0.386     & 69.06  & 81.63 & 83.59 \\
Study     & 41.08   & 0.987   & 0.026   & 96.86 & 97.88 & 98.86 & Room\_1     & 34.72   & 0.931  & 0.134   & 78.38  & 81.41 & 92.11 & 0113\_00    & 31.19   & 0.878   & 0.320     & 98.59  & 98.12 & 98.67 \\
Office    & 46.38   & 0.996   & 0.006   & 97.83 & 96.87 & 100.0 & Room\_2     & 37.32   & 0.963  & 0.115   & 77.58  & 80.33 & 84.78 & 0192\_00    & 28.19   & 0.732   & 0.376     & 96.95  & 98.24 & 99.40 \\ \hline
Average   & 44.17   & 0.992   & 0.013   & 96.87 & 97.42 & \textbf{99.80} & Average     & 36.13   & 0.949  & 0.116   & 75.68 & 77.14 & \textbf{81.03} & Average     & 28.21   & 0.784   & 0.383     & 86.25  & 89.71& \textbf{95.60} 
\end{tabular}
}
\vspace{-.25cm}
\caption{Quantitative results on three datasets. The metric for object decomposition is AP$^{0.75}$.} 
\label{tbl:decomp_dm-sr}
\vspace{-.3cm}
\end{table*}

\begin{figure}[bht]
\centering
   \includegraphics[width=.95\linewidth]{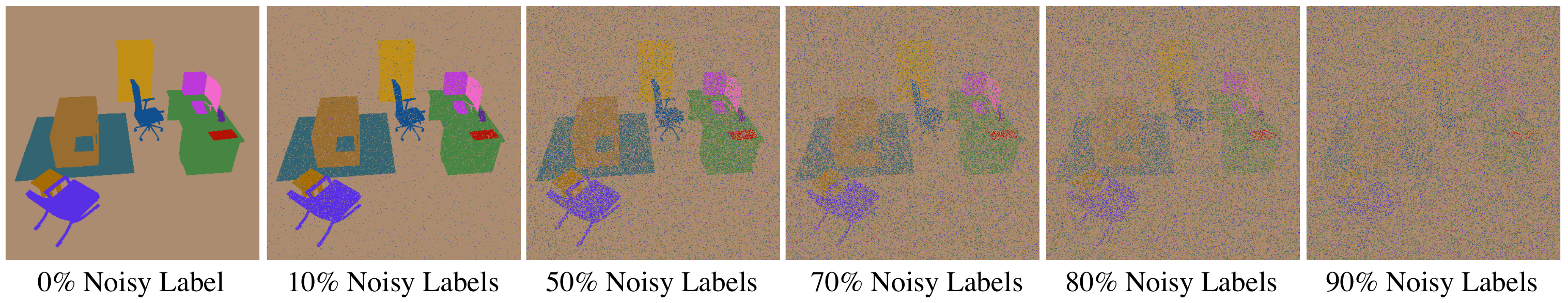}
   \vspace{-0.3cm}
\caption{Examples of 2D object labels with different levels of noise.}
   \vspace{-0.3cm}
\label{fig:exp_noisyLabels}
\end{figure}

\textbf{Robustness to Noisy 2D Labels}: Since our method inherently has multi-view consistency while the 2D segmentation methods do not, it is expected that our method has better robustness to inaccurate and noisy 2D labels in training. To validate this advantage, we conduct the following experiments on DM-SR dataset. Particularly, as illustrated in Figure {\ref{fig:exp_noisyLabels}}, we randomly assign incorrect object labels to different amounts of image pixels of all training images (10\%/50\%/70\%/80\%/90\%), and then our method and baselines are all trained with these noisy 2D labels.

\textbf{Analysis}: As shown in Table {\ref{tbl:robust_dm-sr}}, it can be seen that our method still achieves an excellent object segmentation score (AP$^{0.75}$ = 74.08) on testing/novel views, even though 80\% of 2D labels are incorrect in training. By contrast, both baselines fail catastrophically once more than 50\% of labels are noisy in training. Basically, this is because our dedicated losses $\ell_{2d\_obj}$/$\ell_{3d\_empty}$/$\ell_{3d\_obj}$ explicitly take into account object geometry consistency across multi-views, thus allowing the estimated 3D object codes to be resistant against wrong labels, whereas the existing 2D object segmentation methods only independently process single images, being easily misled by wrong labels in training. This clearly demonstrates the remarkable robustness of our method. More experiments and quantitative/qualitative results regarding robustness are in Appendix.

\textbf{Extension to Panoptic Segmentation:} Our method can be easily extended to tackling panoptic segmentation by adding an extra semantic branch parallel to object code branch. Due to the limited space, quantitative and qualitative results are provided in Appendix.

\vspace{-.2cm}
\begin{table*}[h]\tabcolsep=0.15cm
\centering
\resizebox{1.\linewidth}{!}{
\begin{tabular}{r|ccc|ccc |ccc |ccc |ccc }
 & \multicolumn{3}{c|}{10\% Noisy Labels}  & \multicolumn{3}{c|}{50\% Noisy Labels} & \multicolumn{3}{c|}{70\% Noisy Labels} & \multicolumn{3}{c|}{80\% Noisy Labels} & \multicolumn{3}{c}{90\% Noisy Labels}   \\ \hline

  & MR & SwinT & \textbf{Ours} &  MR & SwinT & \textbf{Ours} &  MR & SwinT & \textbf{Ours} &  MR & SwinT & \textbf{Ours} &  MR & SwinT & \textbf{Ours}  \\
Bathroom  & 98.93 & 98.96 & 99.81   &54.52 & 62.94 & 99.63   &1.86  & 7.71  & 99.02  & 1.89  & 7.78  & 58.09  & 1.79  & 7.11  & 9.69 \\
Bedroom   & 98.61 & 98.85 & 100.0   &75.41 & 92.94 & 100.0   &3.00  & 3.96  & 100.0  & 2.96  & 3.18  & 82.83  & 2.95  & 2.75  & 4.25 \\
Dinning   & 95.57 & 97.88 & 98.41   &46.52 & 43.50 & 85.48   &1.40  & 1.05  & 81.91  & 1.51  & 1.09  & 63.50  & 1.36  & 0.96  & 14.44 \\
Kitchen   & 98.82 & 98.81 & 100.0   &84.93 & 93.60 & 100.0   &4.09  & 8.42  & 100.0  & 4.02  & 5.04  & 51.87  & 4.04  & 4.69  & 1.80 \\
Reception & 80.96 & 91.00 & 100.0   &32.78 & 42.11 & 100.0   &1.03  & 6.27  & 100.0  & 0.73  & 1.96  & 100.0  & 0.22  & 1.64  & 37.63 \\
Rest      & 93.58 & 94.50 & 99.64   &52.75 & 51.99 & 99.32   &1.63  & 1.57  & 99.33  & 2.46  & 2.02  & 66.74  & 1.63  & 1.69  & 11.11\\
Study     & 93.07 & 97.94 & 98.58   &49.03 & 60.04 & 97.97   &1.44  & 6.24  & 98.03  & 1.50  & 6.06  & 72.62  & 1.40  & 5.48  & 40.72 \\
Office    & 95.09 & 97.00 & 100.0   &66.28 & 69.11 & 100.0   &4.13  & 2.96  & 100.0  & 2.61  & 3.94  & 97.00  & 2.66  & 2.48  & 0.00 \\ \hline
Average   & 94.33&96.87&\textbf{99.56}&57.78&64.53&\textbf{97.80}  &2.32  & 4.77  & \textbf{97.29} & 2.21  & 3.88  & \textbf{74.08} & 2.00  & 3.35  & \textbf{14.96} 
\end{tabular}
}
\vspace{-.3cm}
\caption{Quantitative object decomposition AP$^{0.75}$ scores on noisy DM-SR dataset.}
\label{tbl:robust_dm-sr}
\end{table*}

\subsection{3D Object Manipulation}\label{sec:exp_obj_mani}
In this section, we directly use our model trained on the images of Group 1 to test on the remaining images of Groups 2/3/4/5 in DM-SR dataset. In particular, with the trained model, we feed the known manipulation information of Groups 2/3/4/5 into Algorithm \ref{alg:inv_query}, generating images and 2D object masks. These (edited) images and masks are compared with the ground truth 2D views. For comparison, we carefully train Point-NeRF models in scene-specific fashion and apply the same manipulation information to its learned point clouds followed by novel view rendering.

\textbf{Analysis}: Table {\ref{tbl:manip_dm-sr}} shows that the quality of our novel view rendering is clearly better than Point-NeRF \citep{Xu2022Point-NeRF:Fields}, although it decreases after manipulation compared with non-manipulation in Table \ref{tbl:decomp_dm-sr}, primarily because the lighting factors are not decomposed and the illumination of edited objects shows discrepancies. However, the object decomposition is still nearly perfect, as also shown for deformation manipulation in Figure \ref{fig:qualitative_deform}. More results are in Appendix.

\subsection{Ablation Study}

To evaluate the effectiveness of our key designs of object field and manipulator, we conduct two ablation studies. 1) We only remove the loss functions $(\ell_{3d\_{empty}} + \ell_{3d\_{obj}})$ which are jointly designed to learn correct codes for empty 3D points, denoted as w/o $\ell_{3d}$. 2) During manipulation, we remove the step ``Tackle visual occlusions" in Algorithm \ref{alg:inv_query}, denoted as w/o VO. As shown in Table \ref{tbl:manip_dm-sr}, both of our loss functions for empty space regularization and visual occlusion handling step are crucial for accurate 3D scene decomposition and manipulation. More ablation results are in Appendix. 

\begin{table*}[t]
\centering
\resizebox{0.95\linewidth}{!}{
\begin{tabular}{c|cccc|cccc}
 & \multicolumn{4}{c|}{Translation}  & \multicolumn{4}{c}{Rotation}   \\
 & PSNR$\uparrow$   & SSIM$\uparrow$   & LPIPS$\downarrow$  & AP$^{0.9}\uparrow$  & PSNR$\uparrow$ & SSIM$\uparrow$  & LPIPS$\downarrow$   & AP$^{0.9}\uparrow$  \\ \hline

Point-NeRF & 25.79 & 0.847 & 0.202 &- & 25.21 & 0.818 & 0.203 &- \\
Ours    & \textbf{33.94} & \textbf{0.975} & \textbf{0.033} & \textbf{89.33} & \textbf{31.94} & \textbf{0.969} & \textbf{0.038} & \textbf{85.68} \\ \hline
Ablation 1: Ours (w/o $\ell_{3d}$) & 32.84    & 0.967   & 0.048   & 87.26   & 30.38   & 0.945    & 0.090  & 82.46   \\
Ablation 2: Ours (w/o VO)          & 33.54     & 0.970    & 0.045   & 86.93   & 30.57   & 0.953    & 0.076   & 82.43  \\ \hline

 & \multicolumn{4}{c|}{Scale}   & \multicolumn{4}{c}{Joint}   \\
  & PSNR$\uparrow$  & SSIM$\uparrow$   & LPIPS$\downarrow$  & AP$^{0.9}\uparrow$    & PSNR$\uparrow$  & SSIM$\uparrow$  & LPIPS$\downarrow$  & AP$^{0.9}\uparrow$  \\ \hline

Point-NeRF & 25.83 & 0.848 & 0.202 &- & 23.51 & 0.816 & 0.226 &- \\
 Ours   & \textbf{33.40} & \textbf{0.971} & \textbf{0.037} & \textbf{86.05} & \textbf{30.65} & \textbf{0.965} & \textbf{0.045} & \textbf{81.70} \\ \hline
Ablation 1: Ours (w/o $\ell_{3d}$)   & 31.84  & 0.959  & 0.062  & 83.31 & 29.95  & 0.947   & 0.088   & 77.36  \\
Ablation 2: Ours (w/o VO)       & 32.43    & 0.964   & 0.054   & 76.33   & 29.85  & 0.951    & 0.075   & 74.71   \\
\end{tabular}
}
\vspace{-0.2cm}
\caption{Quantitative results of object manipulation and ablation studies on DM-SR dataset.}
\label{tbl:manip_dm-sr}
\end{table*}

 \section{Discussion and Conclusion}
 
\vspace{-0.3cm}
We have shown that it is feasible to simultaneously reconstruct, decompose, manipulate and render complex 3D scenes in a single pipeline only from 2D views. By adding an object field component into the implicit representation, we successfully decompose all individual objects in 3D space. The decomposed object shapes can be further freely edited by our visual occlusion aware manipulator. 

One limitation is the lack of decomposing lighting factors, which is non-trivial and left for future work. In addition, manipulation of originally occluded objects or parts may produce artifacts due to the inaccuracy of learned radiance fields, although these artifacts can be easily repaired by applying simple heuristics such as continuity of surfaces or using better neural backbones.

\clearpage

\textbf{Acknowledgements:} This work was supported in part by National Natural Science Foundation of China under Grant 62271431, in part by Shenzhen Science and Technology Innovation Commission under Grant JCYJ20210324120603011, in part by Research Grants Council of Hong Kong under Grants 25207822 \& 15225522.

\bibliography{references}
\bibliographystyle{iclr2023_conference}

\appendix
\clearpage
\section{Additional Implementation Details}
\begin{figure*}[ht]
\centering
   \includegraphics[width=1.\linewidth]{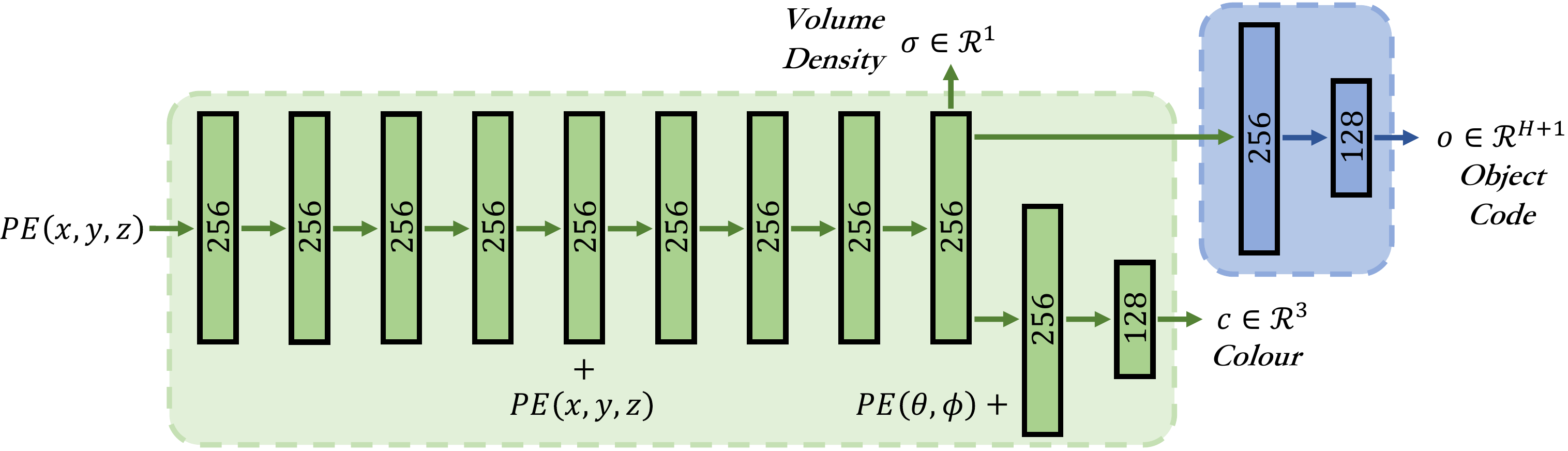}
   \vspace{-.6cm}
\caption{\nickname{} architecture. The positional encoding $PE(\cdot)$ of the location $(x, y, z)$ and viewing direction $(\theta, \phi)$ are taken as the inputs of our network. The volume density $\sigma$ and object code $\boldsymbol{o}$ are the functions of the location while the colour additionally depends on the viewing direction.}
\label{fig:architecture_details}
\vspace{-.6cm}
\end{figure*}

\paragraph{\textbf{Network Architecture}} 
The detailed architecture of our simple pipeline is shown in Figure~\ref{fig:architecture_details}.
\vspace{-0.2cm}

\paragraph{\textbf{2D Object Matching and Supervision}} 
As illustrated in Figure~\ref{fig:dm-nerf_2dsup}, assuming we generate $L$ images of 2D object predictions $\{I_1 \dots I_l \dots I_L \}, I_l \in \mathbf{R}^{U \times V \times (H + 1)}$ and have the paired $L$ images of 2D ground truth object labels $\{\bar{I}_1 \dots \bar{I}_l \dots \bar{I}_L \}, \bar{I}_l \in \mathbf{R}^{U \times V \times T}$, in which $H$ is the predefined number of objects and $T$ represents the number of ground truth objects. 

For each pair, we firstly take the first $H$ solid object predictions of $I$ and reshape it to $M \in \mathbf{R}^{N \times H}$, where $N=U \times V$. Likewise, $\bar{I}$ is reshaped to $\bar{M} \in \mathbf{R}^{N \times T}$. Then, $M$ and $\bar{M}$ are fed into Hungarian algorithm \citep{Kuhn1955} to associate every ground truth 2D object mask with a unique predicted 2D object mask according to Soft Intersection-over-Union (sIoU) and Cross-Entropy Score (CES) \citep{Yang2019d}. Formally, the Soft Intersection-over-Union (sIoU) cost between the $h^{th}$ predicted box and the $t^{th}$ ground truth box in the $\l^{th}$ pair is defined as follows: 
\vspace{-0.1cm}
\begin{equation}
\setlength{\belowdisplayskip}{-1pt}
\boldsymbol{C}^{sIoU}_{h,t} = \frac{ \sum_{n=1}^N(M^n_h * \bar{M}^n_t)}{ \sum_{n=1}^N M^n_h+\sum_{n=1}^{N}\bar{M}^n_t- \sum_{n=1}^{N}(M^n_h *\bar{M}^n_t)}
\vspace{0.2cm}
\end{equation}
where $M^n_h$ and $\bar{M}^n_t$ are the $n^{th}$ values of $M_h$ and $\bar{M}_t$. In addition, we also consider the cross-entropy score between $M_h$ and $\bar{M}_t$ which is formally defined as:
\vspace{-0.1cm}
\begin{equation}
\setlength{\belowdisplayskip}{-2pt}
\boldsymbol{C}^{CES}_{h,t} =-\frac{1}{N} \sum_{n=1}^{N} \left[\bar{M}^n_t\log M^n_h + (1 - \bar{M}^n_t)\log(1-M^n_h)\right]
\end{equation}

After association, we reorder the predicted object masks to align with the $T$ ground truth masks, and then we directly minimize the cost values of all ground truth objects in every pair of images.
\vspace{-0.15cm}
\begin{equation}
    sIoU_l = \frac{1}{T}\sum_{t=1}^T(\boldsymbol{C}^{sIoU}_{t,t}) \quad CES_l = \frac{1}{T}\sum_{t=1}^T(\boldsymbol{C}^{CES}_{t,t})
\vspace{-0.1cm}
\end{equation}

\paragraph{\textbf{Training Details}}
For all experiments, we set the batch size as 3072 rays just to fully use the memory. 
For each ray, we sample 64 points and 128 additional points in the coarse and fine volume, respectively. The Adam optimizer with default hyper-parameters ($\beta_1=0.9$, $\beta_2=0.999$, and $\epsilon=10^{-7}$) is exploited. The learning rate is set to $5 \times 10^{-4}$ and decays exponentially to $5 \times 10^{-5}$ over the course of optimization. The optimization for a single scene typically take around 200--300k iterations to converge on a single NVIDIA RTX3090 GPU (about 17--25 hours).
\vspace{-0.2cm}
\paragraph{\textbf{Evaluation Details}}
We use the Mask-RCNN code open-sourced by the Matterport at \url{https://github.com/matterport/Mask_RCNN} and follow their procedure to calculate the AP values. Note that, we regard IoU values as scores during the ranking procedure.  
\vspace{-0.2cm}
\paragraph{\textbf{Adaptation of Point-NeRF for Object Manipulation}}
During training, to ensure the quality of Point-NeRF on our scene-level DM-SR dataset, we directly use the ground truth dense point cloud (400 million points per scene) instead of the MVSNet generated one when generating the neural point cloud. During the inference of manipulation, we firstly feed the spatial locations of all neural points into our DM-NeRF to infer the corresponding object codes. After determining the points to be edited, we follow the pre-defined manipulation information to transform the corresponding neural points to desired locations. Then, the new neural point cloud is used to render an image with object manipulation from a given view, by point-based volume rendering in Point-NeRF.

\vspace{-0.1cm}
\section{Additional Dataset Details}
\vspace{-0.1cm}
To quantitatively evaluate geometry manipulation, we create a \textbf{s}ynthetic dataset with 8 types of different and complex indoor \textbf{r}ooms (shown in Figure ~\ref{fig:app_dm-sr}), called DM-SR, containing path-traced images that exhibit complicated geometry. The room types and designs follow Hypersim dataset \citep{Roberts2021} and the rendering trajectories follow NeRF synthetic dataset \citep{Mildenhall2020}. Each scene has a physical size of $\sim$12$\times$12$\times$3 meters. Overall, we firstly create and render 8 static scenes, and then manipulate each scene followed by second round rendering. 

In our new dataset, 13 common classes of objects (chair, desk, television, fridge, bathtub, etc.) are introduced. The raw object meshes are downloaded from \url{https://free3d.com/}. We apply different scale/pose transformations on objects and then compose eight common indoor rooms, including bathroom, dinning room, restroom, etc.. We follow the default world coordinate system in Blender: the positive x, y and z axes pointing right, forward and up, respectively. 

To increase realism, we set various types of textures and environment lights for different objects and scenes. Four commonly used types of lights (point, sun, spot, area) are included and the strength is limited within 1000W. During rendering, a camera with 50 degrees field of view is added to generate RGB, depth, semantic and instance images at the resolution of 400$\times$400 pixels. For each scene, the training RGB images are rendered from viewpoints randomly sampled on the upper hemisphere. The viewpoints of testing images are sampled following a smooth spiral trajectory. In addition, instance images are generated by the ray cast function built in Blender.

\begin{figure*}[hb]
\centering
   \includegraphics[width=.85\linewidth]{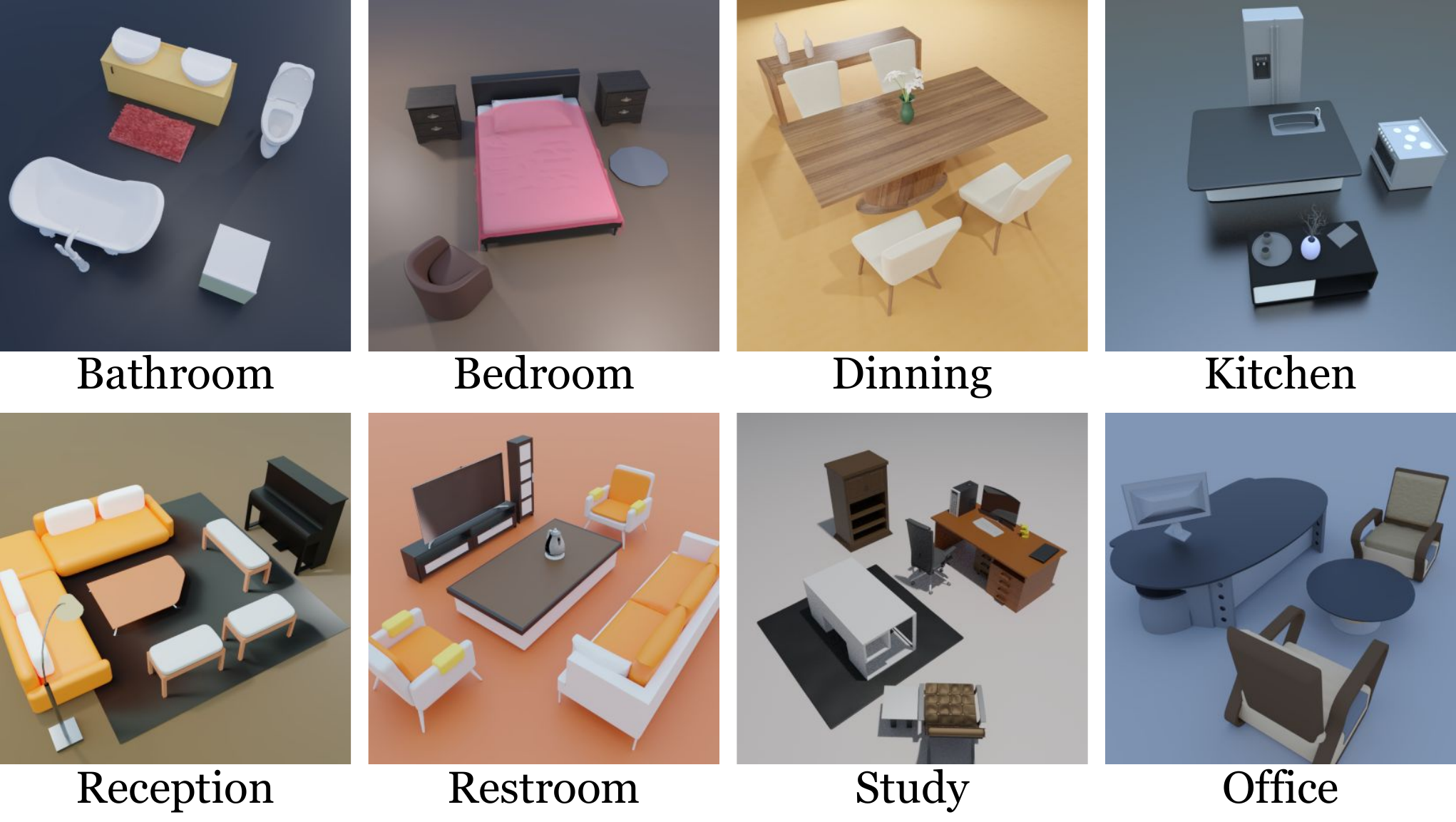}
   \vspace{-0.25cm}
\caption{Eight different indoor scenes of our DM-SR dataset. }
   \vspace{-0.25cm}
\label{fig:app_dm-sr}
\end{figure*}

\section{Additional Quantitative Results}

\paragraph{\textbf{Ablations of MaskRCNN}}
We provide additional quantitative results of AP scores with IoU thresholds at 0.5 and 0.75 in Table~\ref{tab:maskrcnn} to compare four groups of experiments for MaskRCNN~\citep{He2017a} on the selected eight scenes of ScanNet. The settings are as follows:
\begin{table*}[ht]
\vspace{-0.3cm}
\centering
     \resizebox{.85\linewidth}{!}{
\begin{tabular}{c|cccc|cccc}
         & \multicolumn{4}{c|}{AP$^{0.50}\uparrow$}         & \multicolumn{4}{c}{AP$^{0.75}\uparrow$}          \\ \hline
         & G\_1 & G\_2     & G\_3 & G\_4        & G\_1 & G\_2 & G\_3     & G\_4        \\ \hline
0010\_00 & 88.34   & 91.77       & 90.58   & 92.80          & 76.30   & 74.51   & 82.61       & 83.90          \\
0012\_00 & 89.89   & 92.68       & 93.63   & 93.49          & 85.76   & 82.77   & 86.35       & 86.90          \\
0024\_00 & 83.45   & 87.15       & 84.26   & 87.18          & 57.46   & 49.38   & 67.08       & 69.87          \\
0033\_00 & 92.81   & 93.94       & 93.69   & 93.74          & 88.42   & 87.59   & 88.93       & 88.70          \\
0038\_00 & 96.98   & 96.99       & 96.94   & 97.01          & 95.92   & 94.88   & 95.87       & 96.01          \\
0088\_00 & 85.01   & 88.07       & 85.95   & 90.04          & 63.29   & 94.88   & 73.34       & 69.06          \\
0113\_00 & 97.97   & 98.12       & 97.60   & 98.59          & 97.60   & 98.59   & 98.17       & 98.59          \\
0192\_00 & 97.78   & 97.79       & 97.78   & 97.94          & 96.76   & 95.79   & 95.26       & 96.95          \\ \hline
Average  & 91.53   & {\ul 93.31} & 92.55   & \textbf{93.85} & 82.69   & 84.80   & {\ul 85.95} & \textbf{86.25}
\end{tabular}
}
\caption{Average scores of MaskRCNN on 8 scenes of ScanNet.}
\vspace{-0.25cm}
\label{tab:maskrcnn}
\end{table*}

\begin{enumerate}[itemsep = -0.0mm, leftmargin= 10 pt]
    \item[$\bullet$] \textbf{G\_1:} Train a single model on the 8 scenes together from scratch, and then evaluate. 
    \item[$\bullet$] \textbf{G\_2:} Load a single pretrained model, and then finetune and evaluate it on 8 scenes together.
    \item[$\bullet$] \textbf{G\_3:} Train 8 models on the 8 scenes separately from scratch, and then evaluate respectively.
    \item[$\bullet$] \textbf{G\_4:} Load 8 copies of the pretrained model, and then finetune and evaluate on 8 scenes separately. 
\end{enumerate}
Table \ref{tab:maskrcnn} shows that Group 4 (G\_4) achieves the highest scores, which also has the fairest experimental settings we can set up for comparison.

\paragraph{\textbf{Ablations of the Object Field}}
Since the backbone of our pipeline is completely the same as the original NeRF, and our proposed object field component is supervised by $\ell_{2d\_{obj}}$ from 2D signals and by $(\ell_{3d\_{empty}} + \ell_{3d\_{obj}})$ from 3D signals. Note that, the losses $(\ell_{3d\_{empty}} + \ell_{3d\_{obj}})$ only involve a single hyper-parameter $\Delta d$ as shown in Equations \ref{eq:scores_3demp}/\ref{eq:loss_3demp}/\ref{eq:loss_3dobj}. To comprehensively evaluate the effectiveness of these components, we conduct additional experiments with the following ablated versions of our object field along with our main experiments. 
\vspace{-0.1cm}
\begin{enumerate}[itemsep = -0.mm, leftmargin= 10 pt]
    \item[$\bullet$] Without $(\ell_{3d\_{empty}} + \ell_{3d\_{obj}})$: These two losses are designed to learn correct codes for empty 3D points. In this case, we optimize the object field component only by $\ell_{2d\_{obj}}$ from 2D signals.
    
    \item[$\bullet$] With $(\ell_{3d\_{empty}} + \ell_{3d\_{obj}})$: We additionally learn correct codes for empty 3D points using such losses but with different $\Delta d$ (0.025/0.05/0.10 meters), denoted as w/0.025, w/0.05, w/0.10. 
\end{enumerate}
\vspace{-0.1cm}
From Tables~\ref{tbl:ablation_dm-sr}/\ref{tbl:ablation_replica}/\ref{tbl:ablation_scannet}, we find that $\Delta d$ = 0.05 achieves better
scene decomposition quality. It is also clear that the pipeline trained without $(\ell_{3d\_{empty}} + \ell_{3d\_{obj}})$ has worse AP scores than the pipeline trained with $(\ell_{3d\_{empty}} + \ell_{3d\_{obj}})$. In fact, different choices of $\Delta d$ produce very close results, showing that our proposed supervision for empty 3D points is rather robust.
\vspace{-0.1cm}
\begin{table*}[hb]
    \centering
     \resizebox{.9\linewidth}{!}{
\begin{tabular}{c|cccc|cccc}
\multirow{2}{*}{\begin{tabular}[c]{@{}c@{}}Synthetic\\ Rooms\end{tabular}} & \multicolumn{4}{c|}{AP$^{0.75}\uparrow$}     & \multicolumn{4}{c}{AP$^{0.90}\uparrow$}      \\ \cline{2-9} 
                                                                           & w/o   & w/0.025      & w/0.05           & w/0.10 & w/o   & w/0.025      & w/0.05           & w/0.10 \\ \hline
Bathroom                                                                   & 100.0 & 100.0       & 100.0          & 100.0 & 98.17 & 97.72       & 98.50          & 98.16 \\
Bedroom                                                                    & 100.0 & 100.0       & 100.0          & 100.0 & 100.0 & 100.0       & 100.0          & 100.0 \\
Dinning                                                                    & 99.49 & 99.55       & 99.66          & 99.49 & 81.87 & 81.77       & 81.72          & 81.91 \\
Kitchen                                                                    & 100.0 & 100.0       & 100.0          & 100.0 & 100.0 & 100.0       & 100.0          & 100.0 \\
Reception                                                                  & 100.0 & 100.0       & 100.0          & 100.0 & 100.0 & 100.0       & 100.0          & 100.0 \\
Rest                                                                       & 99.88 & 99.89       & 99.89          & 99.89 & 98.77 & 98.97       & 99.03          & 99.03 \\
Study                                                                      & 98.77 & 98.81       & 98.86          & 98.71 & 92.19 & 92.31       & 92.15          & 92.12 \\
Office                                                                     & 100.0 & 100.0       & 100.0          & 100.0 & 100.0 & 100.0       & 100.0          & 100.0 \\ \hline
Average                                                                    & 99.77 & {\ul 99.78} & \textbf{99.80} & 99.76 & 96.38 & 96.35       & \textbf{96.43} & {\ul 96.40}
\end{tabular}
}
    \caption{Quantitative results of scene decomposition from our method on DM-SR dataset. AP scores with IoU thresholds at 0.75 and 0.90 are reported.}
	\label{tbl:ablation_dm-sr}
\end{table*}

\begin{table*}[hb]
    \centering
     \resizebox{.9\linewidth}{!}{
\begin{tabular}{c|cccc|cccc}
\multirow{2}{*}{\begin{tabular}[c]{@{}c@{}}Reconstructed\\ Rooms\end{tabular}} & \multicolumn{4}{c|}{AP$^{0.75}\uparrow$}     & \multicolumn{4}{c}{AP$^{0.90}\uparrow$}      \\ \cline{2-9} 
                                                                               & w/o   & w/0.025      & w/0.05           & w/0.10       & w/o   & w/0.025      & w/0.05           & w/0.10 \\ \hline
Office\_0                                                                      & 79.41 & 75.10       & 82.71          & 76.43       & 57.80 & 56.85       & 55.07          & 53.71 \\
Office\_2                                                                      & 82.77 & 80.83       & 81.12          & 83.70       & 64.68 & 65.02       & 68.34          & 61.18 \\
Office\_3                                                                      & 67.38 & 78.08       & 76.30          & 75.07       & 48.53 & 54.77       & 55.90          & 53.79 \\
Office\_4                                                                      & 65.51 & 64.72       & 70.33          & 73.94       & 47.17 & 50.13       & 53.68          & 53.60 \\
Room\_0                                                                        & 77.60 & 79.73       & 79.83          & 76.03       & 51.21 & 49.63       & 49.35          & 46.69 \\
Room\_1                                                                        & 87.63 & 88.85       & 92.11          & 86.14       & 69.20 & 67.78       & 74.21          & 63.32 \\
Room\_2                                                                        & 84.01 & 84.69       & 84.78          & 83.29       & 58.01 & 62.69       & 62.83          & 57.70 \\ \hline
Average                                                                        & 77.76 & 78.86       & \textbf{81.03} & {\ul 79.23} & 56.66 & {\ul 58.12} & \textbf{59.91} & 55.71
\end{tabular}
}
    \caption{Quantitative results of scene decomposition from our method on Replica dataset. AP scores with IoU thresholds at 0.75 and 0.90 are reported.}
	\label{tbl:ablation_replica}
\end{table*}

\begin{table*}[t]
    \centering
     \resizebox{.9\linewidth}{!}{
\begin{tabular}{c|cccc|cccc}
\multirow{2}{*}{\begin{tabular}[c]{@{}c@{}}Real-world\\ Rooms\end{tabular}} & \multicolumn{4}{c|}{AP$^{0.75}\uparrow$}           & \multicolumn{4}{c}{AP$^{0.90}\uparrow$}      \\ \cline{2-9} 
                                                                            & w/o   & w/0.025      & w/0.05           & w/0.10       & w/o   & w/0.025      & w/0.05           & w/0.10 \\ \hline
0010\_00                                                                    & 95.03 & 94.13       & 94.82          & 94.62       & 86.29 & 86.72       & 90.45          & 86.51 \\
0012\_00                                                                    & 99.25 & 99.25       & 98.86          & 99.75       & 94.40 & 95.71       & 95.35          & 94.54 \\
0024\_00                                                                    & 78.98 & 84.04       & 93.25          & 78.09       & 56.72 & 58.08       & 72.01          & 53.20 \\
0033\_00                                                                    & 94.94 & 92.92       & 97.02          & 95.77       & 89.94 & 87.77       & 93.32          & 88.83 \\
0038\_00                                                                    & 96.57 & 97.02       & 99.17          & 96.58       & 93.43 & 92.96       & 97.58          & 93.82 \\
0088\_00                                                                    & 83.22 & 82.59       & 83.59          & 78.59       & 61.42 & 58.23       & 59.23          & 61.34 \\
0113\_00                                                                    & 92.69 & 92.84       & 98.67          & 98.67       & 85.40 & 85.42       & 96.61          & 96.61 \\
0192\_00                                                                    & 97.60 & 97.60       & 99.40          & 99.80       & 96.48 & 96.44       & 98.32          & 98.88 \\ \hline
Average                                                                     & 92.29 & 92.55       & \textbf{95.60} & {\ul 92.74} & 83.01 & 82.66       & \textbf{87.86} & {\ul 84.22}
\end{tabular}
}
    \caption{Quantitative results of scene decomposition from our method on ScanNet dataset. AP scores with IoU thresholds at 0.75 and 0.90 are reported.}
	\label{tbl:ablation_scannet}
\end{table*}

\paragraph{\textbf{Ablations of Object Field Training Strategy}}
To comprehensively evaluate the object field training strategy, we conduct additional experiments with the following ablated versions of training strategy along with our main experiments. 
\begin{enumerate}[itemsep = -0.mm, leftmargin= 10 pt]
    \item[$\bullet$] Without $Detach$ (w/o): The gradients from object field component can backpropagate to the backbone of NeRF. In this case, the object field and the rendering parts will influence each other.
    \item[$\bullet$]With $Detach$ (w/ ): The object field component only depends on the output representation of NeRF and will not affect the rendering part at all. 
\end{enumerate}

From Tables~\ref{tbl:strategy_dm-sr}/\ref{tbl:strategy_replica}/\ref{tbl:strategy_scannet}, we find the training strategy with $Detach$ achieves better scene rendering and decomposition quality in general. It is clear that the rendering quality drops significantly if our object field component is trained without $Detach$. In contrast, when our object field component is trained with $Detach$, better scene rendering quality and comparable decomposition quality are obtained. In this paper, the training strategy with $Detach$ is adopted.

\begin{table*}[!h]
    \centering
     \resizebox{.85\linewidth}{!}{
\begin{tabular}{c|cc|cc|cc|cc}
          & \multicolumn{2}{c|}{PSNR$\uparrow$} & \multicolumn{2}{c|}{LPIPS$\uparrow$} & \multicolumn{2}{c|}{SSIM$\downarrow$} & \multicolumn{2}{c}{AP$^{0.75}\uparrow$} \\
Detach    & w/                    & w/o         & w/                    & w/o          & w/                    & w/o         & w/                      & w/o           \\ \hline
Bathroom  & 44.05                 & 32.05       & 0.994                 & 0.944        & 0.009                 & 0.150       & 100.0                   & 100.0         \\
Bedroom   & 48.07                 & 32.70       & 0.996                 & 0.927        & 0.009                 & 0.255       & 100.0                   & 100.0         \\
Dinning   & 42.34                 & 33.47       & 0.984                 & 0.895        & 0.028                 & 0.191       & 99.66                   & 99.29         \\
Kitchen   & 46.06                 & 28.49       & 0.994                 & 0.904        & 0.014                 & 0.221       & 100.0                   & 100.0         \\
Reception & 42.59                 & 29.91       & 0.993                 & 0.922        & 0.008                 & 0.190       & 100.0                   & 99.47         \\
Rest      & 42.80                 & 31.33       & 0.994                 & 0.930        & 0.007                 & 0.145       & 99.89                   & 99.47         \\
Study     & 41.08                 & 32.08       & 0.987                 & 0.935        & 0.026                 & 0.161       & 98.86                   & 98.88         \\
Office    & 46.38                 & 32.17       & 0.996                 & 0.935        & 0.006                 & 0.162       & 100.0                   & 100.0         \\ \hline
Average   & \textbf{44.17}        & 31.53       & \textbf{0.992}        & 0.924        & \textbf{0.013}        & 0.185       & \textbf{99.80}          & 99.71        
\end{tabular}
}
    \caption{Quantitative results of our method on DM-SR dataset.}
	\label{tbl:strategy_dm-sr}
\end{table*}

\begin{table*}[!h]
    \centering
     \resizebox{.85\linewidth}{!}{
\begin{tabular}{c|cc|cc|cc|cc}
          & \multicolumn{2}{c|}{PSNR$\uparrow$} & \multicolumn{2}{c|}{LPIPS$\uparrow$} & \multicolumn{2}{c|}{SSIM$\downarrow$} & \multicolumn{2}{c}{AP$^{0.75}\uparrow$} \\
Detach    & w/                    & w/o         & w/                    & w/o          & w/                     & w/o          & w/             & w/o                    \\ \hline
Office\_0 & 40.66                 & 28.20       & 0.972                 & 0.781        & 0.070                  & 0.422        & 82.71          & 82.95                  \\
Office\_2 & 36.98                 & 27.98       & 0.964                 & 0.837        & 0.115                  & 0.361        & 81.12          & 81.69                  \\
Office\_3 & 35.34                 & 26.68       & 0.955                 & 0.817        & 0.078                  & 0.366        & 76.30          & 72.63                  \\
Office\_4 & 32.95                 & 27.19       & 0.921                 & 0.804        & 0.172                  & 0.363        & 70.33          & 77.34                  \\
Room\_0   & 34.97                 & 25.18       & 0.940                 & 0.682        & 0.127                  & 0.403        & 79.83          & 82.26                  \\
Room\_1   & 34.72                 & 26.54       & 0.931                 & 0.717        & 0.134                  & 0.425        & 92.11          & 93.71                  \\
Room\_2   & 37.32                 & 27.43       & 0.963                 & 0.786        & 0.115                  & 0.392        & 84.78          & 83.21                 \\ \hline
Average   & \textbf{36.13}        & 27.03       & \textbf{0.949}        & 0.775        & \textbf{0.116}         & 0.390        & 81.03          & \textbf{81.97}        
\end{tabular}
}
    \caption{Quantitative results of our method on Replica dataset.}
	\label{tbl:strategy_replica}
\end{table*}

\begin{table*}[!h]
    \centering
     \resizebox{.85\linewidth}{!}{
\begin{tabular}{c|cc|cc|cc|cc}
         & \multicolumn{2}{c|}{PSNR$\uparrow$} & \multicolumn{2}{c|}{LPIPS$\uparrow$} & \multicolumn{2}{c|}{SSIM$\downarrow$} & \multicolumn{2}{c}{AP$^{0.75}\uparrow$} \\
Detach   & w/                    & w/o         & w/                    & w/o          & w/                     & w/o          & w/             & w/o                    \\ \hline
0010\_00 & 26.82                 & 22.30       & 0.809                 & 0.697        & 0.381                  & 0.513        & 94.82          & 97.44                  \\
0012\_00 & 29.28                 & 22.98       & 0.753                 & 0.601        & 0.389                  & 0.546        & 98.86          & 97.67                  \\
0024\_00 & 23.68                 & 19.41       & 0.705                 & 0.552        & 0.452                  & 0.573        & 93.25          & 90.45                  \\
0033\_00 & 27.76                 & 22.39       & 0.856                 & 0.743        & 0.342                  & 0.470        & 97.02          & 97.48                  \\
0038\_00 & 29.36                 & 24.79       & 0.716                 & 0.614        & 0.415                  & 0.573        & 99.17          & 98.42                  \\
0088\_00 & 29.37                 & 23.87       & 0.825                 & 0.692        & 0.386                  & 0.513        & 83.59          & 85.45                  \\
0113\_00 & 31.19                 & 22.93       & 0.878                 & 0.727        & 0.320                  & 0.498        & 98.67          & 99.00                  \\
0192\_00 & 28.19                 & 21.97       & 0.732                 & 0.576        & 0.376                  & 0.549        & 99.40          & 98.75                  \\ \hline
Average  & \textbf{28.21}        & 22.58       & \textbf{0.784}        & 0.650        & \textbf{0.383}         & 0.529        & \textbf{95.60} & 95.58        
\end{tabular}
}
    \caption{Quantitative results of our method on ScanNet dataset.}
	\label{tbl:strategy_scannet}
\end{table*}

\paragraph{\textbf{Scene Manipulation and Decomposition}}
To better demonstrate the superiority of our \nickname{} that simultaneously reconstructs, decomposes, manipulates and renders complex 3D scenes in a single pipeline, we conduct additional experiments with the following comparison of scene decomposition along with our main experiments. 
\begin{enumerate}[itemsep = -0.mm, leftmargin= 10 pt]
    \item[$\bullet$] Decomposition after Manipulation: We manipulate objects within a scene and generate corresponding object masks for decomposition using Mask-RCNN with the weights trained on the same scene before manipulation.
    \item[$\bullet$] Simultaneous Manipulation and Decomposition: We appeal to our \nickname{} to simultaneously manipulate and decompose a scene. The weights we used are directly from the same scene but without manipulation.
\end{enumerate}

From Table~\ref{tbl:decom_mani}, we can see that, for the same scene, the AP scores reported by Mask-RCNN~\citep{He2017a} has an obvious decrease after manipulation. However, our method presents very close AP scores for all scenes before and after manipulation. Fundamentally, this is because Mask-RCNN only considers every 2D image independently for object segmentation, while our \nickname{} explicitly leverages the consistency between 3D and 2D across multiple views.

\begin{table*}[hb]
    \centering
     \resizebox{.95\linewidth}{!}{
\begin{tabular}{c|cccc|cccc}
Metric       & \multicolumn{4}{c|}{AP$^{0.5}\uparrow$}                         & \multicolumn{4}{c}{AP$^{0.75}\uparrow$} \\ \cline{1-9}
Method       & \multicolumn{2}{c}{Mask-RCNN} & \multicolumn{2}{c|}{\textbf{Ours}} & \multicolumn{2}{c}{Mask-RCNN} & \multicolumn{2}{c}{\textbf{Ours}} \\
Manipulation & Before         & After        & Before           & After        & Before         & After        & Before             & After     \\ \cline{1-9}
Bathroom     & 97.90          & 96.36        & 100.0            & 99.38        & 93.81          & 88.89        & 100.0              & 97.57     \\
Bedroom      & 98.91          & 97.14        & 100.0            & 100.0        & 97.92          & 94.84        & 100.0              & 99.38     \\
Dinning      & 98.85          & 98.20        & 100.0            & 99.15        & 98.85          & 96.33        & 99.66              & 97.14     \\
Kitchen      & 92.06          & 93.56        & 100.0            & 100.0        & 92.04          & 91.39        & 100.0              & 98.75     \\
Reception    & 98.81          & 97.03        & 100.0            & 100.0        & 98.81          & 94.63        & 100.0              & 99.40  \\
Rest         & 98.89          & 97.18        & 100.0            & 100.0        & 98.89          & 95.86        & 99.89              & 99.86     \\
Study        & 96.87          & 97.64        & 99.69            & 99.41        & 96.86          & 95.75        & 98.86              & 98.38     \\
Office       & 98.93          & 89.97        & 100.0            & 100.0        & 97.83          & 74.24        & 100.0              & 75.94     \\ \cline{1-9}
Average      & 97.65          & 95.88        & \textbf{99.96}   & {\ul 99.74}  & 96.87          & 91.49        & \textbf{99.80}     & {\ul 95.80} 
\end{tabular}
}
    \caption{Quantitative results of scene decomposition from our method and Mask-RCNN on DM-SR dataset. AP scores with IoU thresholds at 0.5 and 0.75 are reported.}
	\label{tbl:decom_mani}
\end{table*}

\paragraph{\textbf{Computation}}
We typically train 200K iterations in $\sim$15 hours on each scene ($\sim$0.27s for each iteration) with the batch size of 3072 rays, which uses $\sim$24GB GPU memory. In contrast, the original NeRF (rendering only) needs $\sim$0.22s for each iteration). During the inference of joint decomposition and rendering, our DM-NeRF costs $\sim$9.3s per image with the batch size of 4096 rays using $\sim$5GB GPU memory. For joint decomposition, manipulation and rendering, $\sim$23.4s are required for each image and $\sim$9GB GPU memory is needed when the batch size is set as 4096 rays. To render an image from a novel view, the original NeRF and Point-NeRF need $\sim$7.8s and $\sim$8.2s, respectively. All training and testing are operated on a single Nvidia GeForce RTX 3090 card.

\clearpage

\section{Additional Experiments for Noisy 2D Labels}

As illustrated in Figure \ref{fig:app_objseg_noisyMR}, to further evaluate the robustness of our method, we use the instance masks estimated by Mask-RCNN as the supervision signals when training our DM-NeRF. Table \ref{tbl:robust_dm-sr_mrcnn} shows the quantitative results on our DM-SR dataset. We can see that even though the 2D labels are inaccurate for training, our method still achieves excellent object decomposition results. 

\begin{table*}[hb]
    \centering
     \resizebox{.95\linewidth}{!}{
\begin{tabular}{ c| c c c c c c c c| c}
  & Bathroom & Bedroom & Dinning & Kitchen & Reception & Rest & Study & Office & Average \\  \hline
 \nickname{} & 95.30 & 97.08 & 95.69 & 94.72 & 99.62 & 98.58 & 97.72 & 98.08 & 97.10 \\   
\end{tabular}
}
    \caption{Quantitative results of scene decomposition from our method trained with noisy 2D labels (estimated by Mask-RCNN) on DM-SR dataset. AP scores with IoU thresholds at 0.75 are reported.}
\label{tbl:robust_dm-sr_mrcnn}
\end{table*}

\section{Extension to Panoptic Segmentation}

\begin{figure*}[hb]
\centering
\includegraphics[width=1.\linewidth]{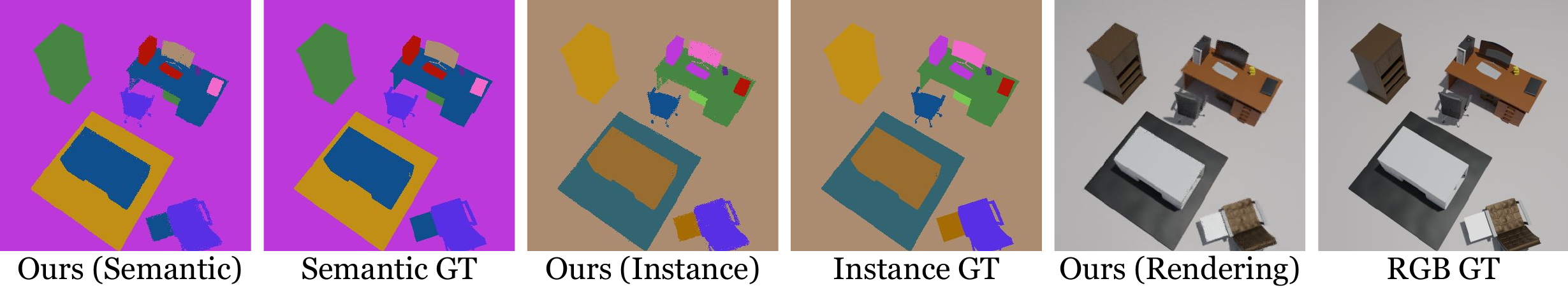}
\vspace{-.4cm}
\caption{Qualitative results of our method for panoptic segmentation on DM-SR dataset.}
\label{fig:panoptic_dm-sr}
\end{figure*}

An extra semantic branch parallel to object code branch is added into our current \nickname{} for panoptic segmentation. Table \ref{tbl:panoptic_dm-sr} shows the quantitative results on our DM-SR dataset where the accurate 2D semantic and instance labels are used to train our network. Figure \ref{fig:panoptic_dm-sr} shows the qualitative results. It can be seen that both semantic categories and object codes are accurately inferred.

\begin{table*}[hb]
    \centering
     \resizebox{.95\linewidth}{!}{
\begin{tabular}{ l| c c c c c c c c| c}
  & Bathroom & Bedroom & Dinning & Kitchen & Reception & Rest & Study & Office & Average \\  \hline
 \nickname{} (Obj: AP$^{0.75}$) & 100.0 & 100.0 & 99.41 & 100.0 & 100.0 & 97.86 & 96.84 & 100.0 & 99.26 \\  
  \nickname{} (Sem: mIoU) & 97.58 & 99.08 & 94.64 & 98.72 & 97.42 & 97.13 & 94.29 & 97.85 & 97.09 \\  
\end{tabular}
}
    \caption{Quantitative results of panoptic segmentation from our method trained with accurate 2D semantic and instance labels on DM-SR dataset. The AP score of all objects and the mIoU score of all categories in each scene are reported respectively.}
\label{tbl:panoptic_dm-sr}
\end{table*}

\section{Additional Qualitative Results}
\begin{figure*}[b]
\centering
   \includegraphics[width=1.\linewidth]{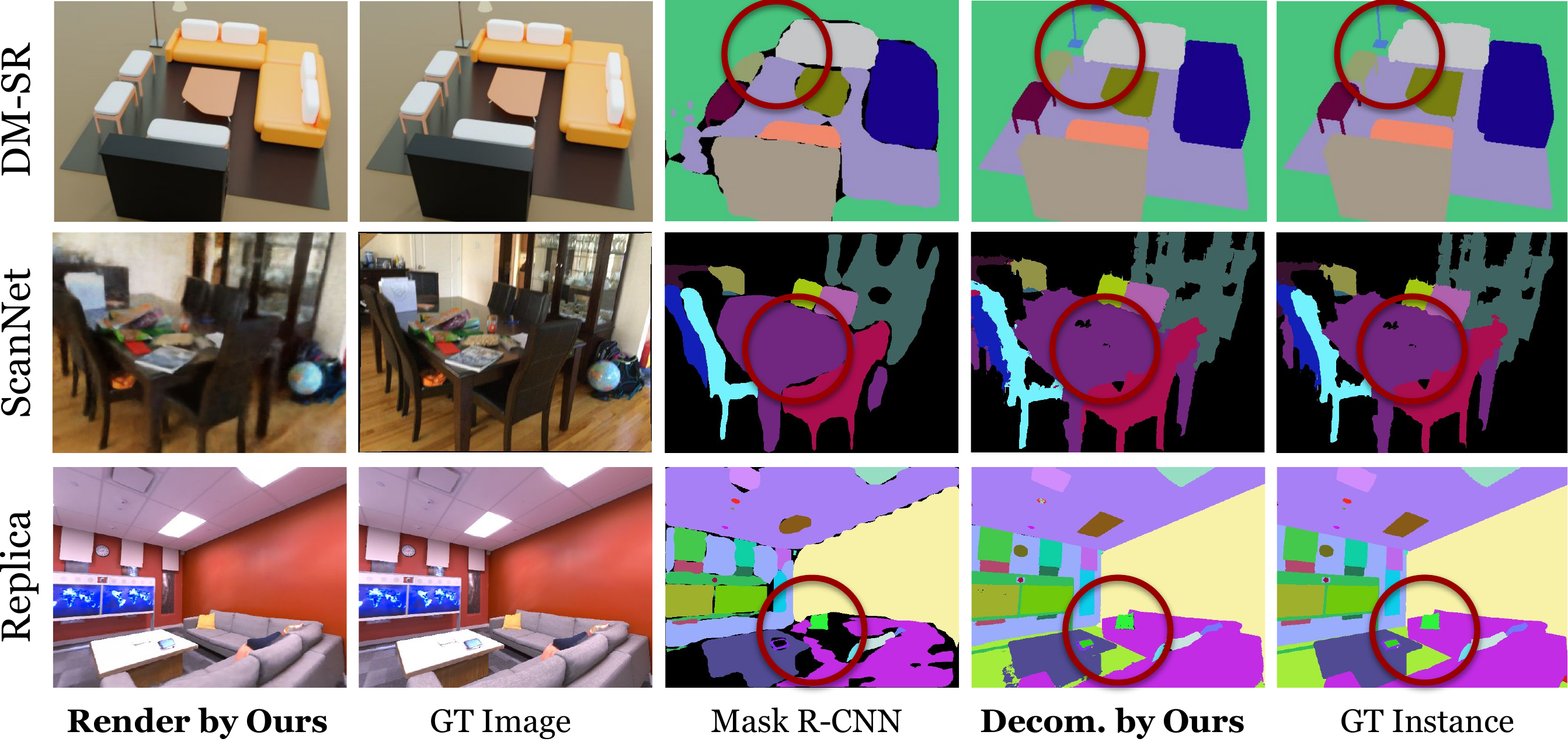}
   \vspace{-0.4cm}
\caption{Qualitative results of our method and the baseline on three datasets: DM-SR, Replica and ScanNet. The dark red circles highlight the differences.}
\label{fig:qualitative_decomp}
\end{figure*}

\begin{figure*}[hb]
\centering
   \includegraphics[width=.95\linewidth]{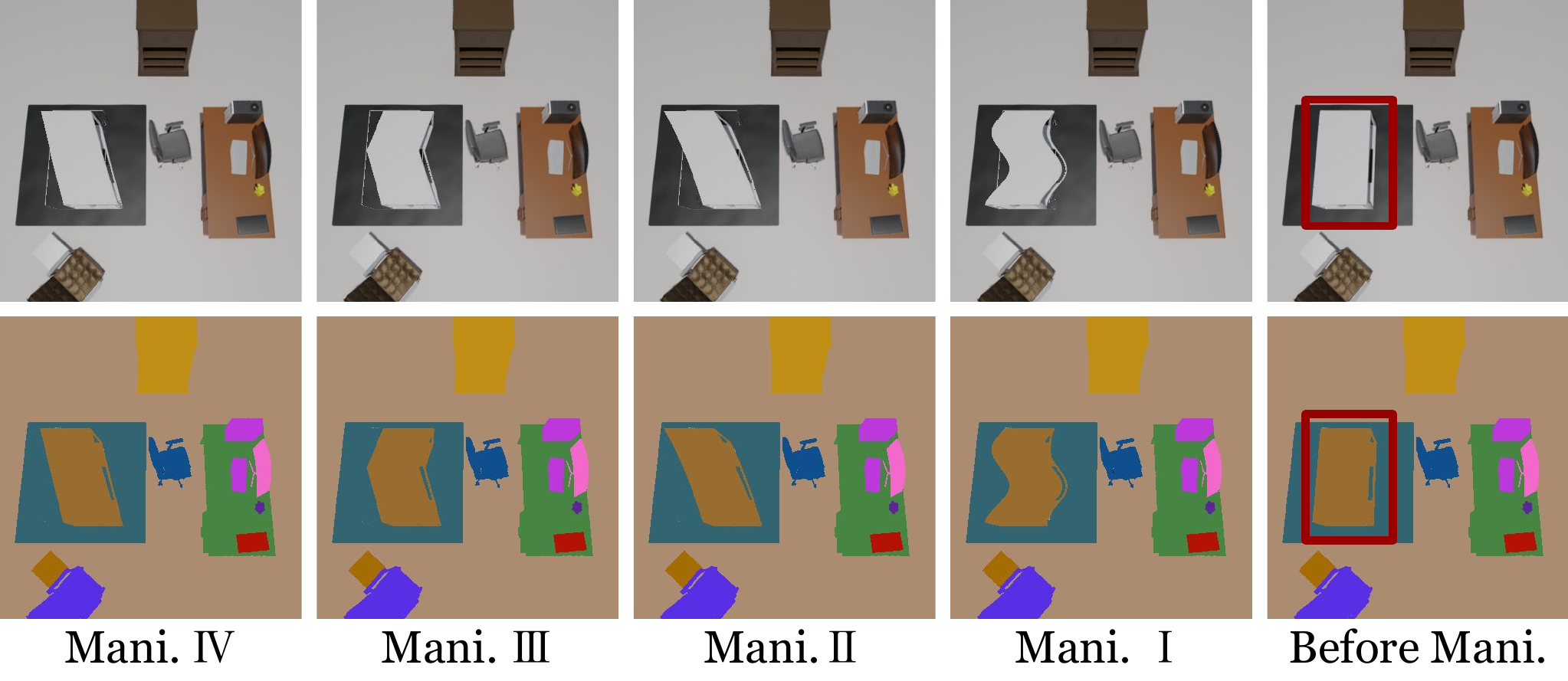}
   \vspace{-0.4cm}
\caption{Qualitative results of our method for object deformation manipulation on DM-SR dataset. The dark red boxes highlight the target table to be manipulated. }
\label{fig:qualitative_deform}
\end{figure*}

\begin{figure*}[hb]
\centering
\includegraphics[width=1.\linewidth]{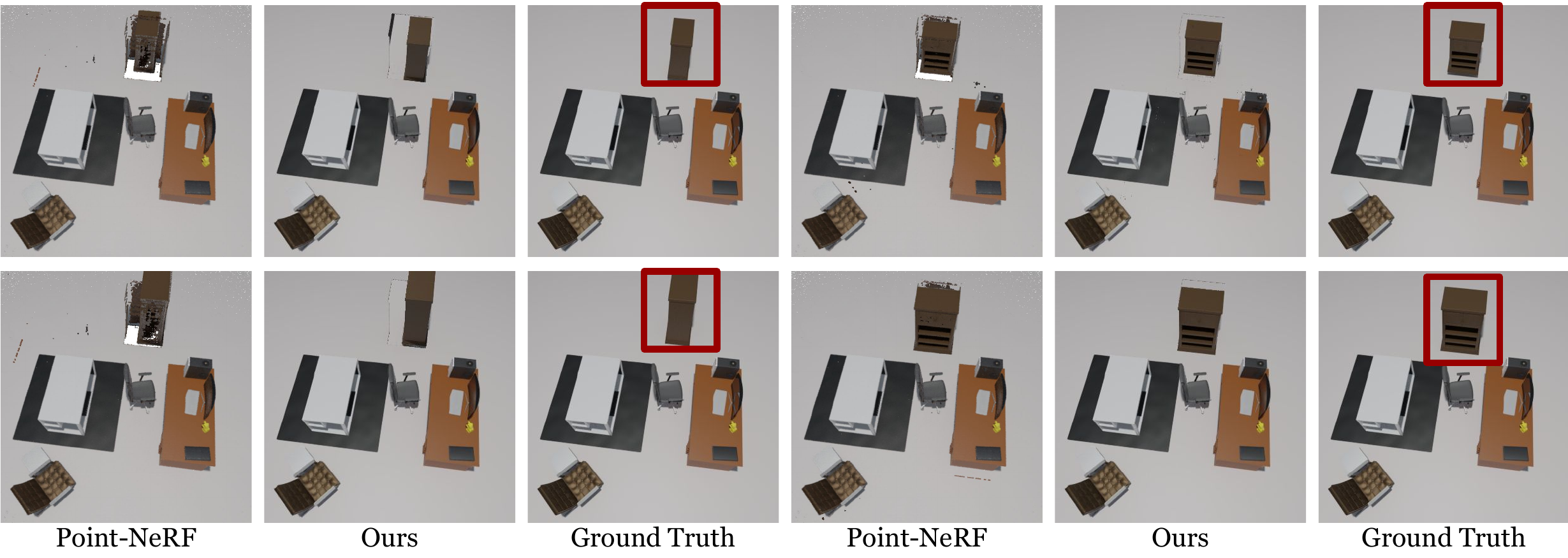}
\vspace{-.4cm}
\caption{Qualitative results of novel view rendering after manipulating 3D objects. It can be seen that our method obtains clearly sharper object shapes after manipulation in 3D space, whereas the baseline Point-NeRF shows obvious artifacts such as holes, primarily because its manipulation is conducted on explicit 3D point clouds followed by neural rendering, but our manipulation is conducted in continuous neural radiance space and therefore has fine-grained results.}
\label{fig:app_objmani_quali}
\end{figure*}

\begin{figure*}[ht]
\centering
   \includegraphics[width=1.\linewidth]{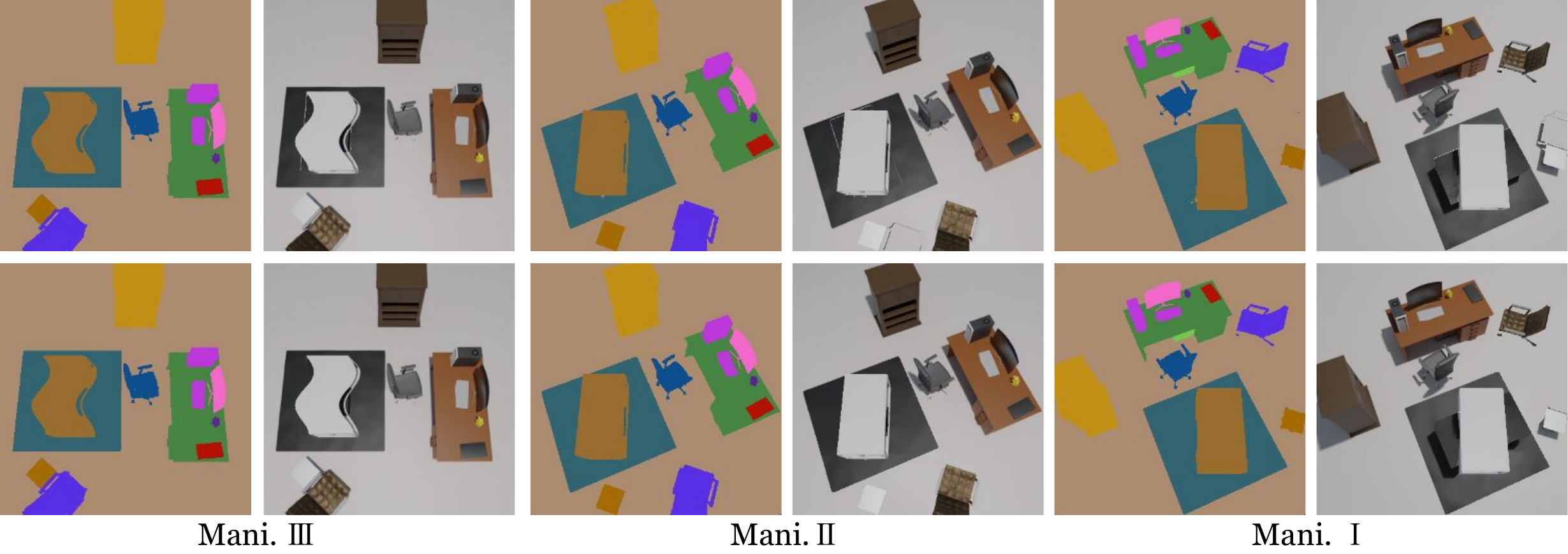}
   \vspace{-.6cm}
\caption{The artifacts can be further mitigated. By introducing a voting strategy, we improve the quality of scene decomposition for the following fine manipulation. To be specific, we also consider 8 neighbouring rays when determining the projected object code $\boldsymbol{\hat{o}}$ along a light ray. If $\boldsymbol{\hat{o}}$ is different from the codes of 8  neighbouring rays and these codes is dominated by one of them by voting (the number of dominated code is greater than 4), then $\boldsymbol{\hat{o}}$ will be reset to the dominated code. Otherwise, the original code will be kept. Such an operation can generate more accurate scene decomposition results to support the Inverse Query Algorithm, especially when visual occlusion happens.}
\vspace{-.1cm}
\label{fig:app_artifacts}
\end{figure*}

\paragraph{\textbf{Scene Decomposition and Manipulation}} Figures \ref{fig:qualitative_decomp}/\ref{fig:qualitative_deform}/\ref{fig:app_objmani_quali}/\ref{fig:app_artifacts} show  qualitative results of 3D scene decomposition and manipulation in Sections \ref{sec:exp_scene_decomp} \& \ref{sec:exp_obj_mani}.

\begin{figure*}[ht]
\centering
\includegraphics[width=1.\linewidth]{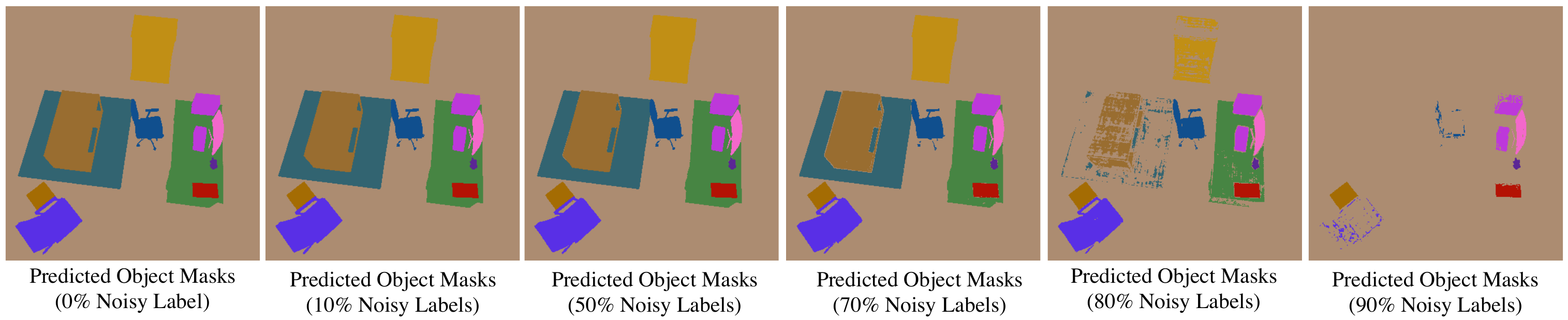}
\vspace{-0.6cm}
\caption{Estimated object masks at novel views from our models trained with different amount of noisy labels. Our method can infer satisfactory results, even though 80\% of 2D labels are incorrect during training, demonstrating the robustness of our method.}
\label{fig:app_objseg_noisy}
\vspace{-.1cm}
\end{figure*}

\begin{figure*}[hb]
\centering
   \includegraphics[width=.7\linewidth]{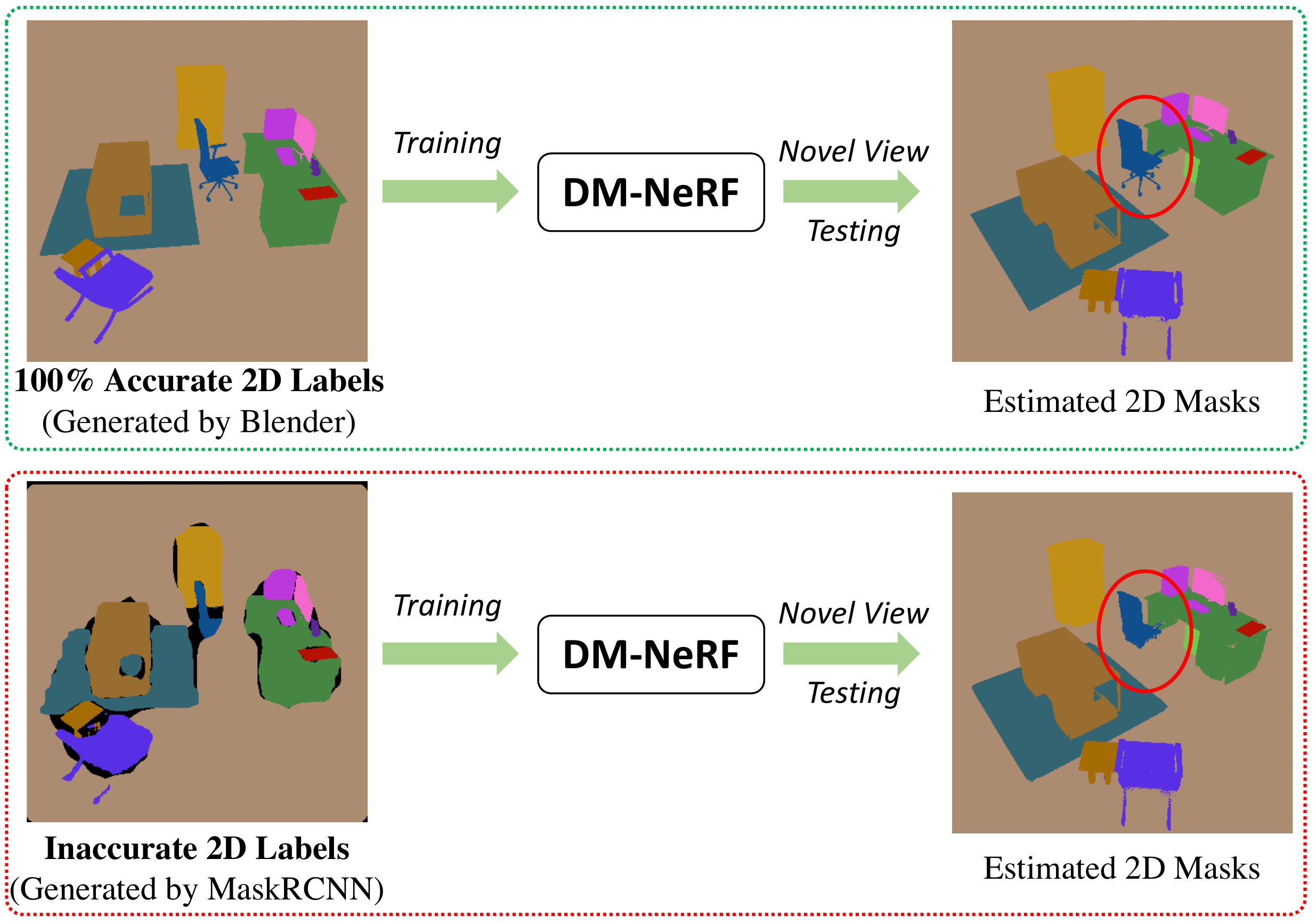}
   \vspace{-0.1cm}
\caption{As shown in the red dotted block, we use the instance masks estimated by Mask-RCNN as the supervision signals when training our DM-NeRF. We can see that even though the 2D labels are inaccurate for training, our method still achieves excellent object decomposition results. The red circles show that only thin chair feet are not segmented using the inaccurate 2D labels in training. This result is consistent with that of Figure \ref{fig:app_objseg_noisy}, showing the remarkable robustness of our method.}
\vspace{-1.6cm}
\label{fig:app_objseg_noisyMR}
\end{figure*}

\paragraph{\textbf{Scene Decomposition}} Figures \ref{fig:app_objseg_noisy}/\ref{fig:app_objseg_noisyMR} shows qualitative results of scene decomposition from our method trained on noisy and inaccurate 2D labels.

\begin{figure*}[t]
\centering
  \includegraphics[width=1.\linewidth]{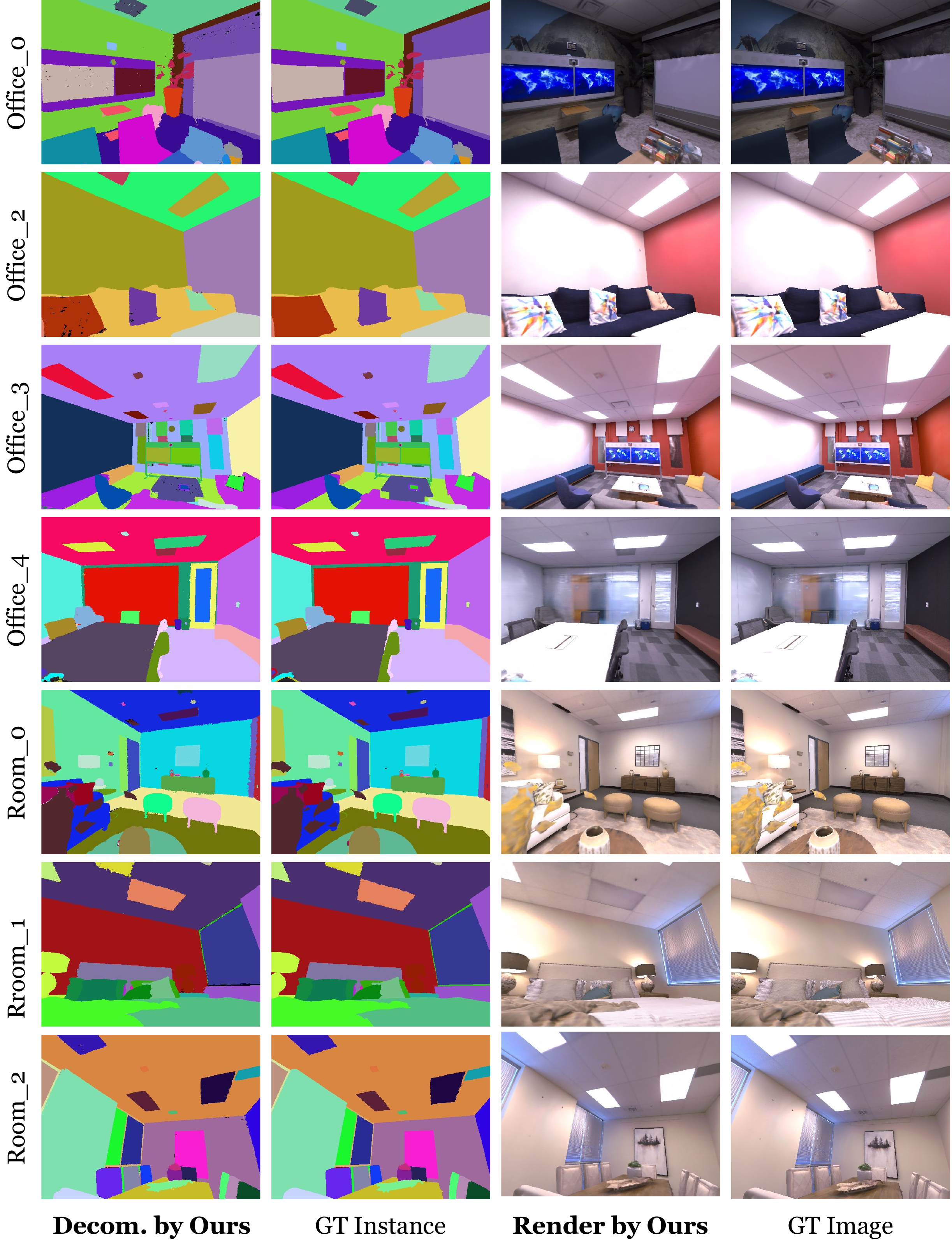}
\caption{Qualitative results for scene rendering and decomposition on Replica.}
\label{fig:app_replica}
\end{figure*}

\begin{figure*}[t]
\centering
  \includegraphics[width=1.\linewidth]{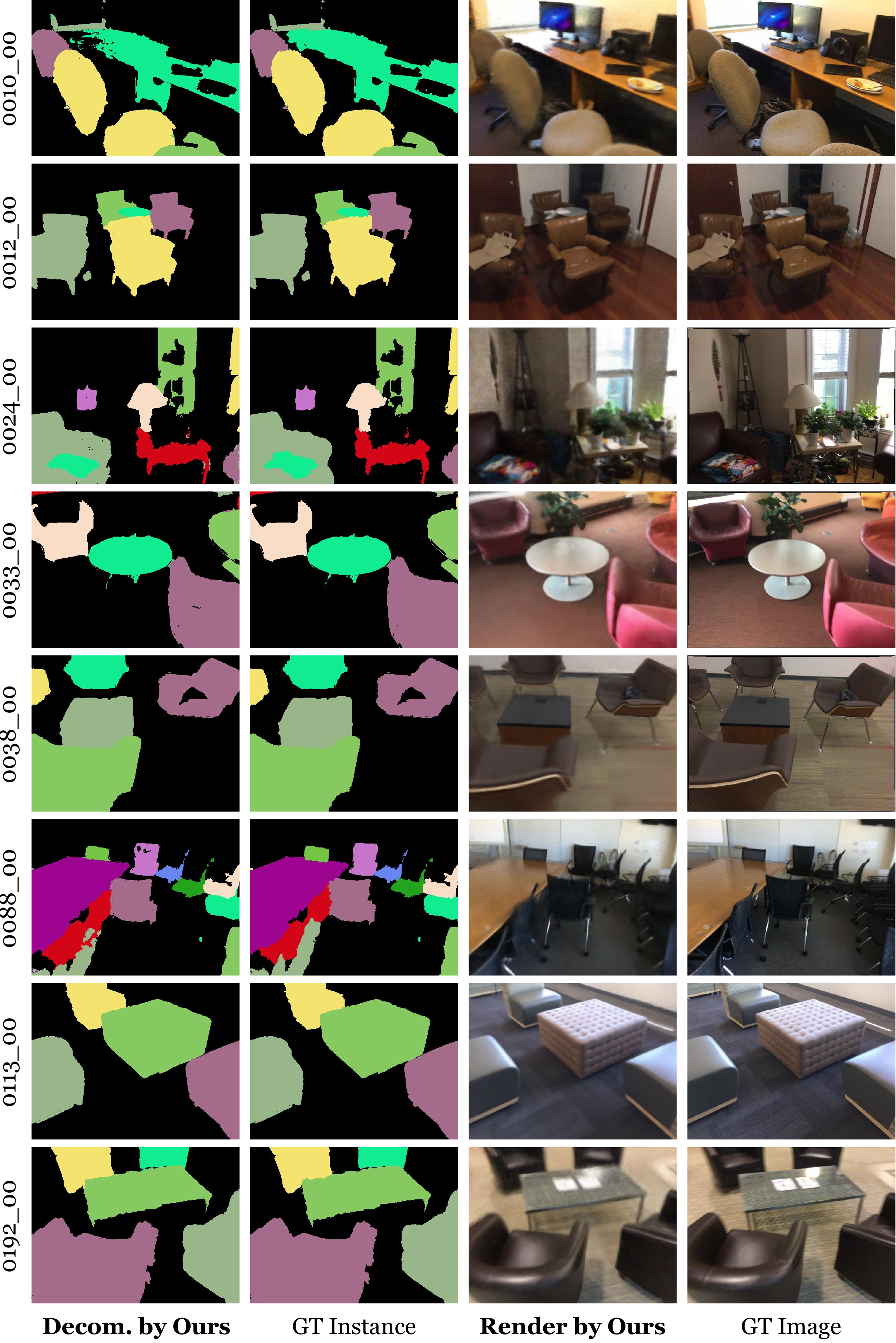}
\caption{Qualitative results for scene rendering and decomposition on ScanNet.}
\label{fig:app_scannet}
\end{figure*}

\paragraph{\textbf{Scene Rendering and Decomposition}}
Figures \ref{fig:app_dm-sr}/\ref{fig:app_replica}/\ref{fig:app_scannet} show additional qualitative results of scene rendering and decomposition.

\begin{figure*}[t]
\centering
   \includegraphics[width=.9\linewidth]{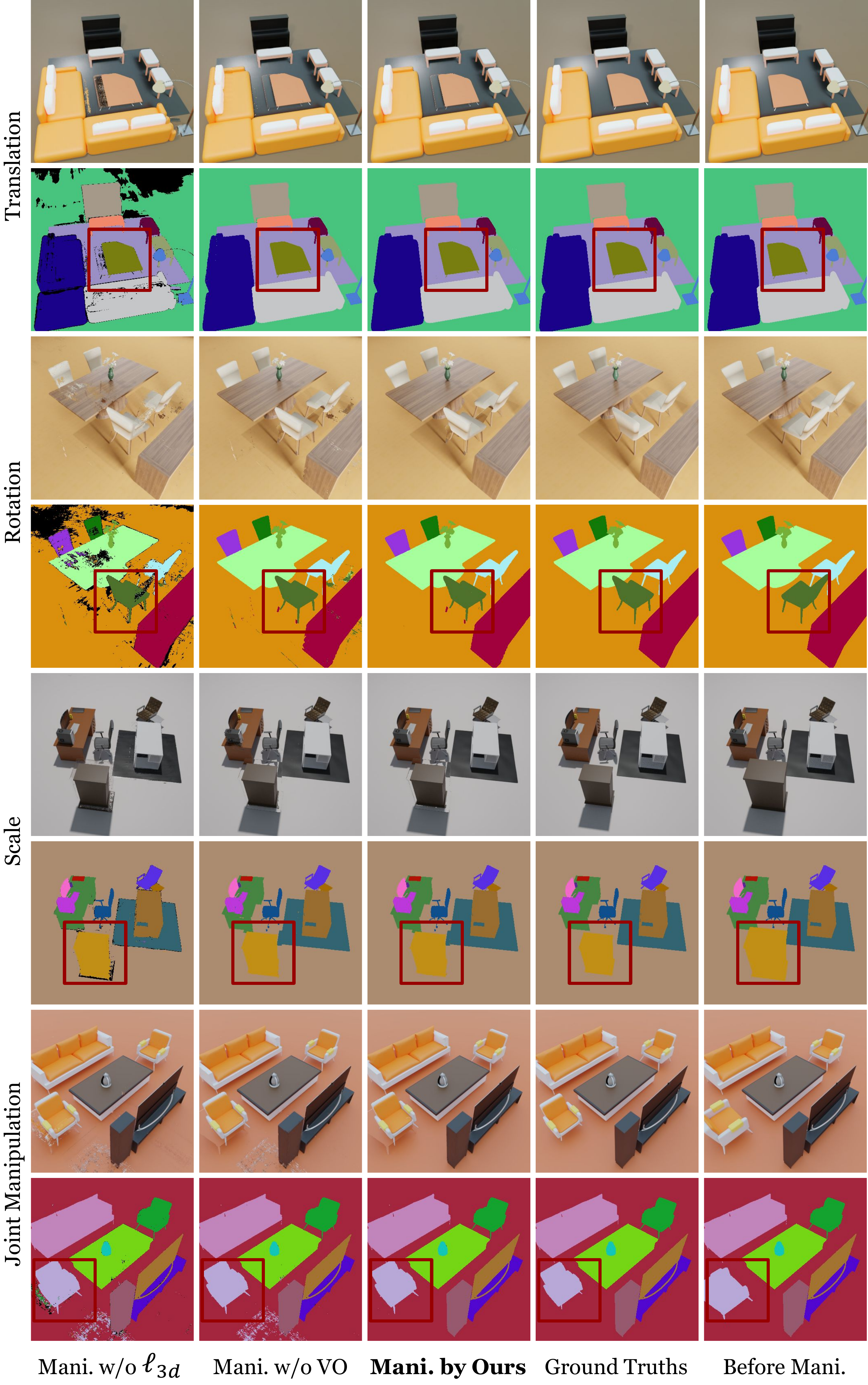}
\caption{Qualitative results of our method for object manipulation on DM-SR dataset. The dark red boxes highlight the differences. }
\label{fig:manip_dm-sr}
\end{figure*}

\begin{figure*}[t]
\centering
  \includegraphics[width=1.\linewidth]{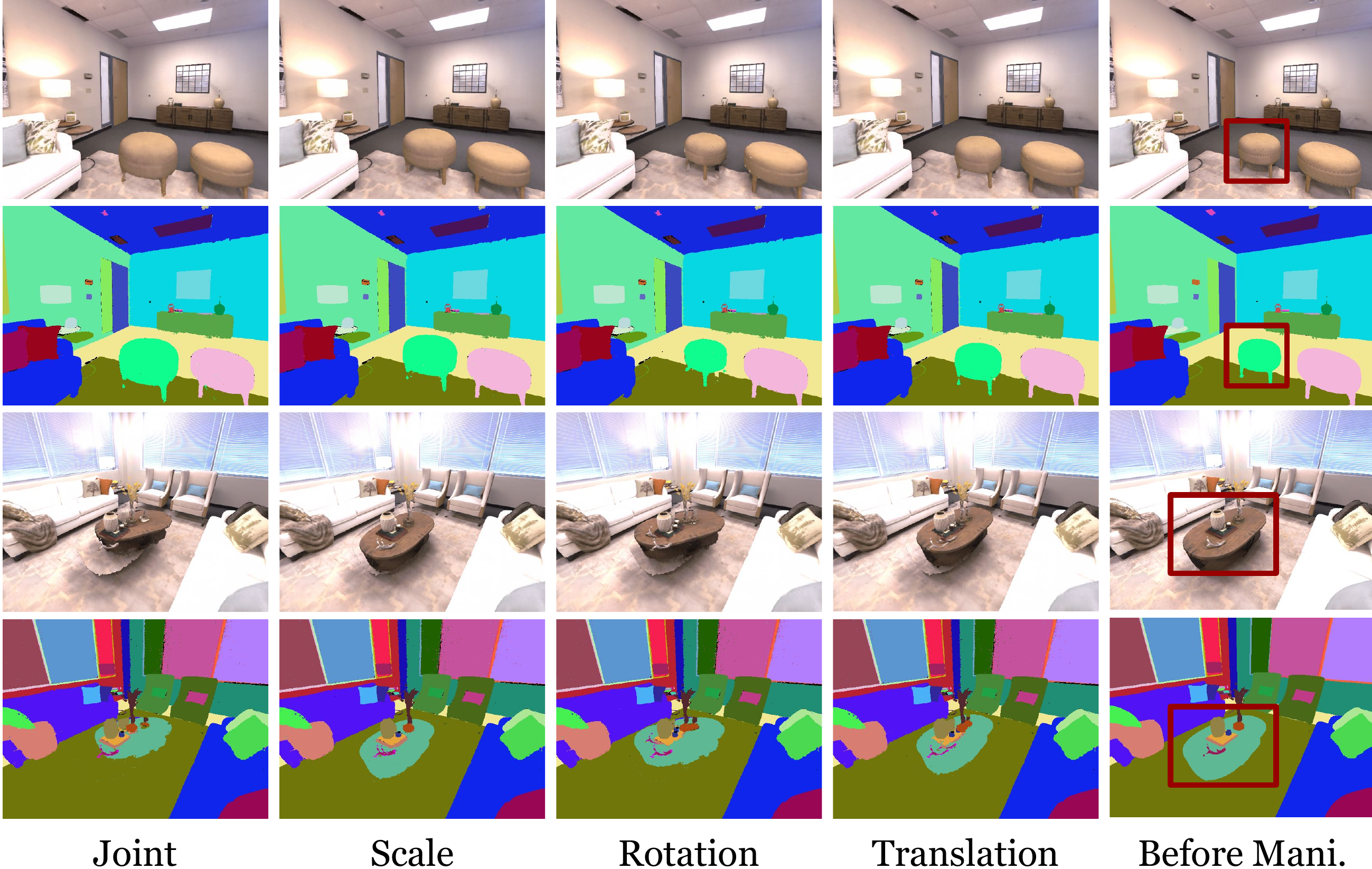}
\caption{Qualitative results for single object manipulation on Replica dataset. The dark red box in the rightmost column highlights the manipulated object.}
\label{fig:app_mani_replica}
\end{figure*}

\begin{figure*}[t]
\centering
  \includegraphics[width=1.\linewidth]{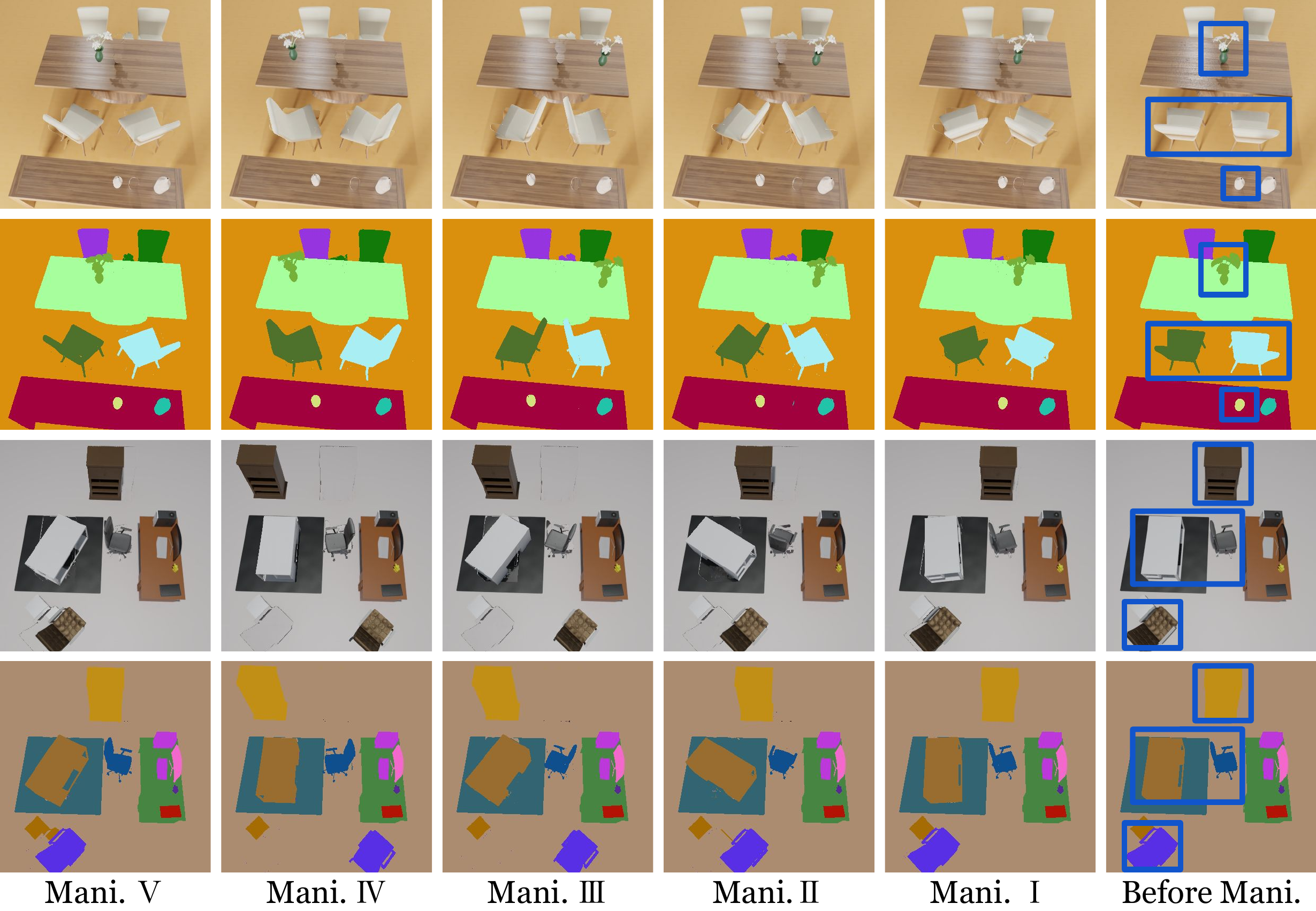}
\caption{Qualitative results for multiple objects manipulation on DM-SR dataset. The dark blue boxes in the rightmost column highlight the manipulated objects.}
\label{fig:app_mani_dm-sr}
\end{figure*}

\paragraph{\textbf{Scene Manipulation}}
Figures \ref{fig:manip_dm-sr}/\ref{fig:app_mani_replica}/\ref{fig:app_mani_dm-sr}/\ref{fig:app_objmani_quali} shows additional qualitative results of scene manipulation with single/multiple objects under various transformations.

\paragraph{More qualitative results can be found in the video available at \url{https://github.com/vLAR-group/DM-NeRF}.}

\clearpage

\clearpage
\begin{figure*}[t]
\centering
  \includegraphics[width=0.8\linewidth]{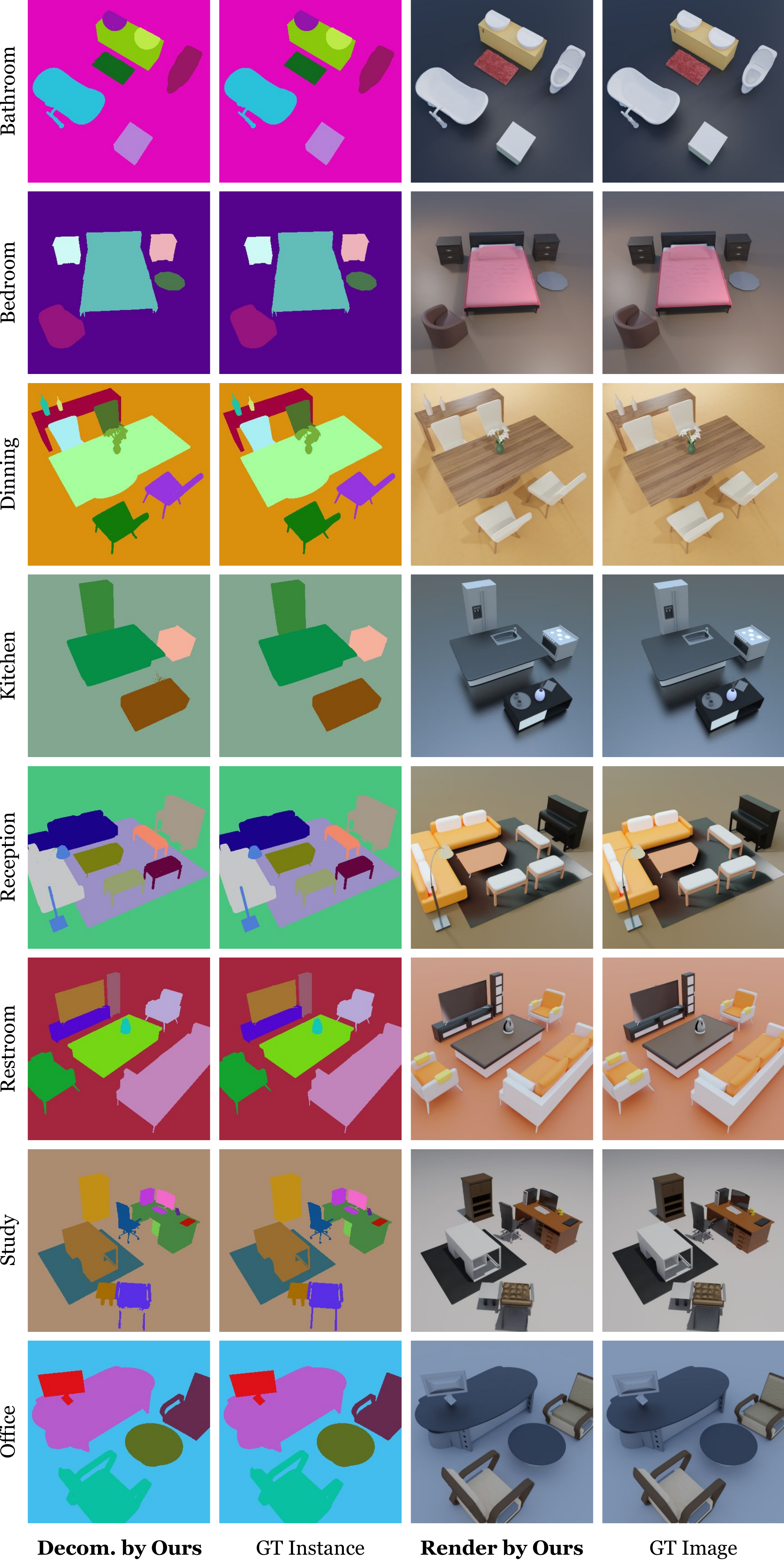}
  \vspace{-0.4cm}
\caption{Qualitative results for scene rendering and decomposition on DM-SR.}
\label{fig:app_dm-sr}
\end{figure*}

\end{document}